%% file: paper.tex
\PassOptionsToPackage{table}{xcolor}
\documentclass[]{sjtu_sai}

\input{math_commands.tex}

\usepackage{pifont}
\usepackage{titletoc}
\usepackage{algorithm}
\usepackage{algorithmic}
\usepackage{enumitem}
\usepackage{wrapfig}
\usepackage{needspace}
\usepackage{array}
\usepackage{colortbl}

\definecolor{darkred}{RGB}{139,0,0}
\definecolor{moongold}{HTML}{B8892F}
\definecolor{costup}{HTML}{BA8157}
\definecolor{dark-gray}{gray}{0.30}
\newcommand{\coreq}[1]{\textcolor{darkred}{\textbf{\textit{#1}}}}
\newcommand{\method}{\textbf{MOON}}
\newcommand{\pub}[1]{{\color{dark-gray}{\scriptsize{[{#1}]}}}}
\newcommand{\cm}{\ding{51}}
\newcommand{\xm}{\ding{55}}
\graphicspath{{fig/}}


\title{%
  \mbox{\textcolor{moongold}{\textbf{MOON}}\,\raisebox{-0.12em}{%
    \includegraphics[height=0.95em]{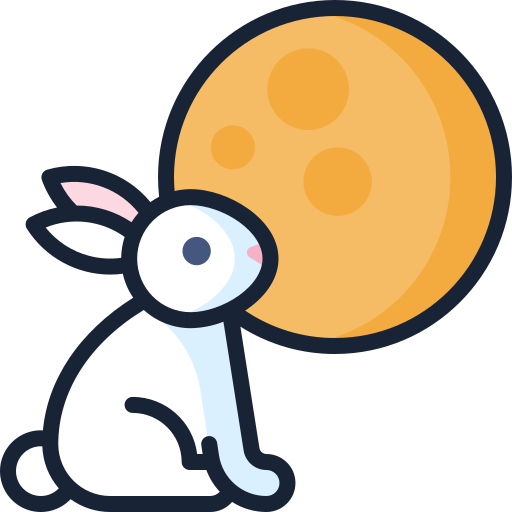}}}:
  Learning When to Forget\\
  and How to \textcolor{moongold}{MO}ve \textcolor{moongold}{O}n\\
  in Vision-and-Language \textcolor{moongold}{N}avigation}

\hypersetup{
  hypertexnames=false,
  pdftitle={MOON: Learning When to Forget and How to Move On in Vision-and-Language Navigation},
  pdfauthor={Jingmeng Zhou, Seson Sun, Binyao Xu, Lingming Zhang, Hanshi Wang, Fudong Ge, Weiming Hu, Jin Gao, Zhipeng Zhang}
}
\leftlogo{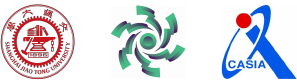}
\input{authors.tex}
\abstract{\input{sec/0.abstract}}

\begin{document}
\maketitle
\input{fig/fig_teaser_moon}
\input{sec/1.introduction}
\input{sec/3.analysis}
\input{sec/4.method}
\input{sec/5.experiments}
\input{sec/6.conclusion}
\input{sec/8.limitations}

\clearpage
\bibliographystyle{plainnat}
\bibliography{references}

\clearpage
\appendix
\input{sec/7.appendix}
\end{document}

%% file: math_commands.tex
\usepackage{amsmath,amsfonts,bm}

\def\eqref#1{equation~\ref{#1}}

\def\1{\bm{1}}

\DeclareMathAlphabet{\mathsfit}{\encodingdefault}{\sfdefault}{m}{sl}
\SetMathAlphabet{\mathsfit}{bold}{\encodingdefault}{\sfdefault}{bx}{n}

\newcommand{\KL}{D_{\mathrm{KL}}}

%% file: authors.tex
\renewcommand\authorformat[2][]{%
  \authorfont{\bfseries\rmfamily #2\ifstrempty{#1}{}{%
    \raisebox{0.6ex}{\normalfont\fontsize{8}{8}\selectfont #1}}}}
\newcommand\authorrow[2][]{%
  \addtolist[#1]{#2}{\authorlist}{\authorformat}{\par}}
\author[1,2,*]{Jingmeng~Zhou}
\author[1,*]{Seson~Sun}
\author[2,*]{Binyao~Xu}
\authorrow{Lingming~Zhang}
\author[1,2]{Hanshi~Wang}
\author[1,2]{Fudong~Ge}
\authorrow[2]{Weiming~Hu}
\author[2,\textdagger]{Jin~Gao}
\author[1,\textdagger]{Zhipeng~Zhang}

\renewcommand\affiliationformat[2][]{{\affiliationfont\itshape\rmfamily
  \ifstrempty{#1}{}{\textsuperscript{\normalfont #1}}#2}}
\renewcommand\affiliation[2][]{%
  \addtolist[#1]{#2}{\affiliationlist}{\affiliationformat}{\par}}
\affiliation[1]{AutoLab, School of Artificial Intelligence, Shanghai Jiao Tong University}
\affiliation[2]{State Key Laboratory of Multimodal Artificial Intelligence Systems, CASIA}

\renewcommand\contributionformat[2][]{{\contributionfont
  \ifstrempty{#1}{}{\textsuperscript{#1}}#2}}
\contribution[*]{Equal contribution}
\contribution[\textdagger]{Corresponding authors.}

%% file: sec/0.abstract.tex
Recent vision-and-language navigation (VLN) policies act on their visual history, and an incorrect action can take them off the instructed route.
Paired interventions on four VLN policies show that clearing the history doubles pooled success at the onset of a deviation but lowers it on route, while later in a deviation, removing only the off-route frames helps much less.
After a deviation, the retained history keeps the policy from turning where a turn is needed, which we call \emph{behavioral inertia}.
We propose \method{}, a framework that equips a frozen VLN policy with two lightweight modules, adding less than $1\%$ of its parameters.
A deviation judge decides when to forget, raising an alarm that clears the history when it predicts that the agent is off route.
A recovery policy, trained with GRPO from alarm states, learns how to move on, turning more of the successes found only by sampling into a single greedy rollout.
Experiments on R2R-CE and RxR-CE show that \method{} reaches the highest R2R-CE success rate among the compared methods and raises the success rate of its frozen base by $5.2$ and $6.0$ points, and an ablation with a random trigger indicates that this gain depends on when the history is cleared.
Without retraining, \method{} also improves zero-shot object-goal navigation on HM3D-OVON.
Together, these results suggest that deviation recovery depends on knowing when to forget, not only how to act.
Code is available at \url{https://github.com/AutoLab-SAI-SJTU/MOON}.

%% file: fig/fig_teaser_moon.tex
\begin{figure}[!htbp]
\centering
\includegraphics[width=\linewidth]{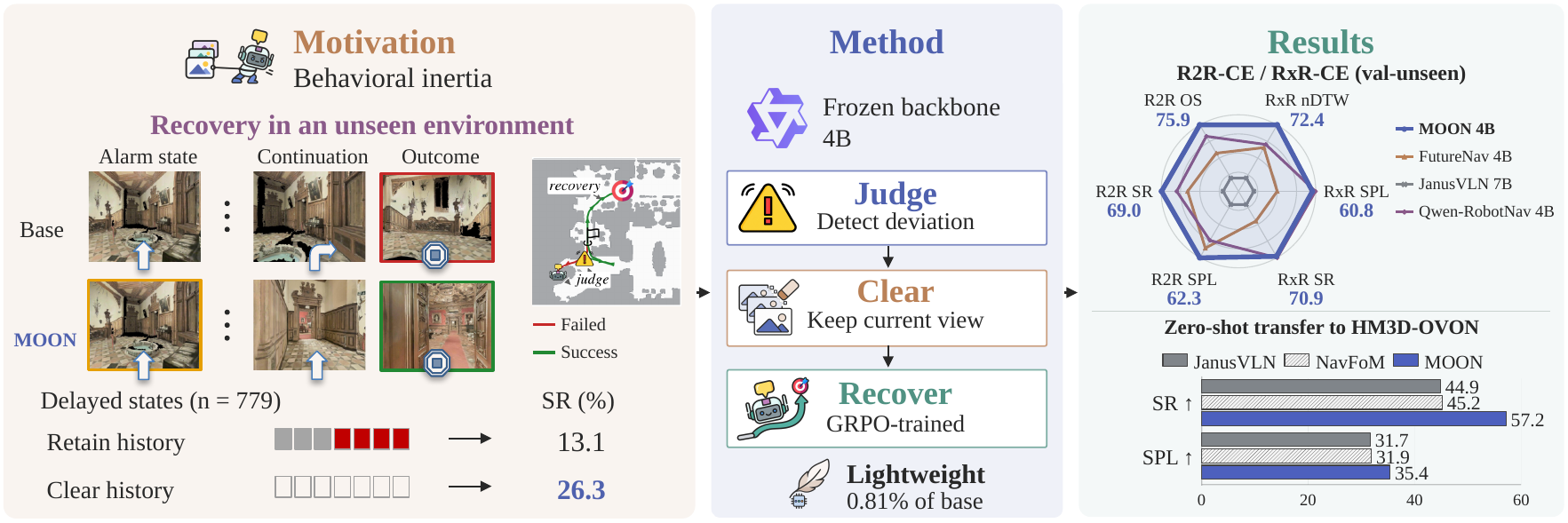}
\caption{\textbf{\method{} learns when to forget and how to move on.}
Left: one paired episode in an unseen environment, in which the base fails and \method{} recovers after the judge's alarm (amber border), and the effect of clearing, which raises pooled success from 13.1\% to 26.3\% across 779 delayed states of four frozen policies.
Middle: a deviation judge triggers history clearing and a trained recovery policy on a frozen 4B backbone.
Right: R2R-CE and RxR-CE val-unseen results against recent leading monocular RGB models of Table~\ref{tab:main}, 4B versions where a model comes in several sizes, and zero-shot HM3D-OVON results against JanusVLN and NavFoM.
Each radar axis is scaled to the range of the plotted models; labeled values are \method{}'s.}
\label{fig:teaser}
\end{figure}

%% file: sec/1.introduction.tex
\section{Introduction}

Vision-and-Language Navigation (VLN) requires an embodied agent to follow natural-language instructions and reach a target location in a 3D environment using egocentric visual observations \citep{anderson2018vln,krantz2020vlnce}.
Recent VLN policies leverage pretrained vision-language models to predict navigation actions from the instruction, the current observation, and the visual history \citep{zhu2026navida,cheng2024navila,wei2025streamvln,zhang2024uninavid}.
Yet long-horizon navigation remains vulnerable to action errors.
An incorrect action can take the agent off the instructed route, causing a \textit{deviation}.
Unless corrected promptly, the agent must continue from an off-route state, where later errors can compound into task failure \citep{ross2011reduction}.
Beyond avoiding deviations, reliable VLN thus requires correcting course and resuming the instructed route after a deviation, which we refer to as \textit{deviation recovery}.

Existing approaches to deviation recovery primarily improve the policy itself.
They expose the policy to off-route states and teach it to return to the instructed route, either by continuing training on corrective trajectories from failed episodes \citep{yu2025correctnav,xin2026decovln} or by reinforcement learning \citep{navgrpo2026}.
These approaches improve overall success, but because they change the policy and evaluate it over complete episodes, which mix on-route and off-route states, they do not isolate the effect of retaining the visual history during recovery.
Appendix~\ref{app:related} reviews related work, including detectors that hand control to a separate controller~\citep{jeong2026navheads}.
Work on human memory suggests that retaining everything is not always useful: forgetting reduces the influence of outdated information on decisions in changing environments \citep{richards2017persistence}, and people segment their experience into a new event when their predictions fail \citep{zacks2007event}.
In VLN, visual history may help before a deviation, when it records progress along the instructed route.
Once the agent deviates, however, newly accumulated observations begin to describe an off-route trajectory that no longer corresponds to the instruction \citep{ke2019tactical}.
The policy must therefore learn to recover while still conditioning on a history increasingly shaped by its own mistakes.
This history could provide useful context for retracing the route or reinforce the deviation.
We ask: \coreq{How does visual history affect recovery after a deviation?}

To answer this question, we run paired interventions in the Habitat simulator \citep{savva2019habitat} on R2R-CE with four VLN policies (NaVIDA, NaVILA, StreamVLN and Uni-NaVid) that realize three kinds of visual memory, without modifying their parameters.
Following prior work \citep{yu2025correctnav,xin2026decovln}, a deviation occurs when the agent's distance from the reference route exceeds a threshold; from each deviation onset, we run the policy twice, once with its accumulated visual history and once with the history cleared, leaving only the instruction and the current frame.
Clearing the history raises the success rate from 13.7\% to 28.6\% across 915 onset states from the four policies and lowers it from 76.0\% to 48.3\% on route; both effects hold for every policy.
A natural explanation for the harm is that the accumulated off-route frames mislead the policy.
We test this explanation at states ten or more steps after the deviation by retaining only the history up to the onset, which removes the later off-route frames while keeping the agent's on-route context.
On these states, doing so raises the success rate only from 13.1\% to 16.3\%, well below the 26.3\% that clearing reaches (Figure~\ref{fig:teaser}, left).
The harm therefore does not come mainly from the off-route frames.

After a deviation, the retained history keeps the policy from turning where the new state calls for a turn.
We call this \emph{behavioral inertia}.
At the deviation onset, after a forward step, a shortest-path expert at the same pose continues without a turn in only 2\% of states.
The four policies continue without a turn through their next decision in 48\% of states, and in 11\% once their history is cleared.
Further interventions on the visual content of a full-length retained history, shuffling it or swapping in one from another scene, lower the success rate on route but barely change it after a deviation.
These findings are consistent with a known failure mode of policies trained on expert demonstrations.
During training, features associated with expert histories reliably predict expert actions, leading the policy to depend on them.
After the policy leaves the expert distribution, these associations may no longer hold, but the policy continues acting on them and compounds its errors \citep{dehaan2019causal,codevilla2019exploring,wen2020fighting,sobolmark2026bpp}.
We hypothesize that, in our setting, the history itself becomes such a feature through its presence and length more than its content: histories filled with current-frame copies or noise still cause most of the harm, and the harm grows with the retained length (Sections~\ref{sec:analysis:history} and~\ref{sec:analysis:inertia}).
Clearing the history reduces this inertia and doubles the recovery success rate.
It does not reveal how to return to the route, however, and most deviations still end in failure.
Yet among onset states where the retained greedy rollout fails, at least one of eight sampled rollouts after clearing succeeds in $61.6\%$, against $30.3\%$ with the history retained (Section~\ref{sec:analysis:recovery}), so there exist many recoveries that a single greedy rollout misses.

Building on these insights, we propose \method{} (Figure~\ref{fig:teaser}, middle), whose two components answer, respectively, when to clear the visual history and how to make recovery succeed after clearing.
Both are added to a frozen policy and together amount to less than 1\% of its parameters.
The first component, a \emph{deviation judge}, clears the visual history when it predicts that the agent is off route, because clearing on route is harmful.
The second component, a \emph{recovery policy}, takes over after clearing.
Because reinforcement learning with verifiable rewards is effective at making behaviors that a policy can already sample more reliable \citep{yue2025does}, we train it with Group Relative Policy Optimization (GRPO) \citep{shao2024deepseekmath} on cleared rollouts from alarm states.

We evaluate \method{} on R2R-CE and RxR-CE (Figure~\ref{fig:teaser}, right).
On R2R-CE it reaches the highest success rate among the compared methods, $2.1$ points above the next best, Qwen-RobotNav-4B~\citep{qwenrobotnav}, and it raises the success rate of its frozen base by $5.2$ points on R2R-CE and $6.0$ on RxR-CE.
A random trigger with as many alarms lowers success below the base, so the gain depends on when the history is cleared.
Without retraining, the same components improve zero-shot open-vocabulary object-goal navigation on HM3D-OVON \citep{yokoyama2024hm3dovon}.
Our contributions are threefold:
\begin{itemize}[leftmargin=1.5em,itemsep=3pt,topsep=3pt,parsep=0pt]
\item \textbf{Analysis.} On four frozen VLN policies, paired interventions show that the visual history is beneficial on route and harmful after a deviation, that removing the off-route frames recovers only a small part of this harm, that the retained history keeps the policy from turning where a turn is needed (behavioral inertia), and that after clearing, sampling finds successes that a single greedy rollout misses (Section~\ref{sec:analysis}).
\item \textbf{Method.} We propose \method{}, two lightweight modules on a frozen policy: a deviation judge that learns when to clear the history and a recovery policy that learns how to recover (Section~\ref{sec:method}).
\item \textbf{Experiments.} \method{} reaches the highest R2R-CE success rate among the compared methods and competitive results on RxR-CE; its gain depends on when the history is cleared, and without retraining it improves zero-shot object-goal navigation (Section~\ref{sec:experiments}).
\end{itemize}

%% file: sec/3.analysis.tex
\section{Motivation and Analysis}\label{sec:analysis}

\subsection{Preliminaries}\label{sec:analysis:prelim}

\paragraph{Problem Formulation.}
We study vision-and-language navigation in continuous 3D environments.
Given a natural-language instruction $I$, at each step $t$ the agent receives an egocentric RGB frame $o_t$ and executes an action $a_t$: move forward $0.25$\,m, turn $15^\circ$ left or right, or stop.
The policy reads its interaction history $H_t = \{(o_s, a_s)\}_{s=0}^{t-1}$ through a memory constructor $M$:
\begin{equation}
a_t \sim \pi_\theta\bigl(\cdot \mid I,\, o_t,\, M(H_t)\bigr).
\label{eq:policy}
\end{equation}
In practice the policies act in chunks: at a \emph{decision point} $t$, $\pi_\theta$ proposes a chunk $c_t$ of atomic actions and executes it before the next decision point, and $t$ always counts atomic steps.
An episode succeeds if the agent stops within the standard success radius of the goal.

\paragraph{Deviation Criterion.}
The simulator provides the agent's position $q_t$ and the reference route $\mathcal{R}$, which the instruction describes and the agent never observes.
Following prior work~\citep{yu2025correctnav,xin2026decovln}, the agent deviates once it is more than $S_d = 2$\,m from the route:
\begin{equation}
h_t = \min_{r\in\mathcal{R}} d_{\mathrm{geo}}(q_t, r),
\qquad t_d = \min\{\,t : h_t > S_d\,\},
\label{eq:deviation}
\end{equation}
where $d_{\mathrm{geo}}$ is the geodesic distance and the deviation onset $t_d$ is $\infty$ for an episode that never deviates.

\begin{figure}[t]
\centering
\captionsetup[sub]{skip=1pt}
\begin{minipage}{0.9\linewidth}
\begin{subfigure}[b]{0.3249\linewidth}
\includegraphics[width=\linewidth]{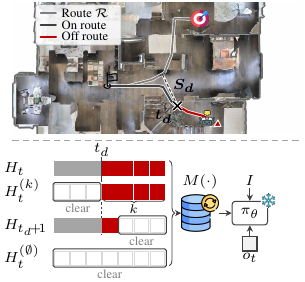}
\caption{History interventions}\label{fig:interventions}
\end{subfigure}\hfill
\begin{subfigure}[b]{0.3249\linewidth}
\includegraphics[width=\linewidth]{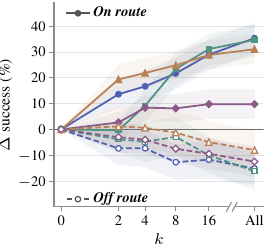}
\caption{History length}\label{fig:history_length}
\end{subfigure}\hfill
\begin{subfigure}[b]{0.3249\linewidth}
\includegraphics[width=\linewidth]{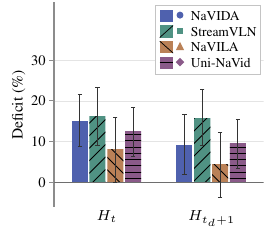}
\caption{History content}\label{fig:history_content}
\end{subfigure}
\end{minipage}
\caption{\textbf{Visual history helps on route but can hinder recovery.} (a)~Boxed steps are cleared. (b,\,c)~Off route: delayed states. Bands and whiskers: $95\%$ paired bootstrap intervals.}
\label{fig:history_response}

\end{figure}

\subsection{Two Opposing Roles of Visual History}\label{sec:analysis:history}

We first ask whether, from the same simulator state, retaining the visual history helps or hurts the policy.
All analyses in this section use the same $600$ randomly sampled R2R-CE val-unseen episodes and four frozen policies that realize $M$ in three ways (Appendix~\ref{app:analysis-details}).
The \emph{full-length retained history} denotes each policy's memory constructed from the complete interaction prefix $H_t$.
\emph{Clearing} replaces $H_t$ with $\emptyset$ while preserving the instruction $I$ and current frame $o_t$.
Each intervention replaces only the history passed to $M$ at a decision point and restarts the policy from the same simulator state with the same remaining step budget (Figure~\ref{fig:interventions}).

\paragraph{Clearing On and Off Route.}
An onset (delayed) state is the first decision point at or after $t_d$ ($t_d+10$).
Pooled rates weight every state of the four policies equally.
Across $915$ paired onset states, clearing raises pooled success from $13.7\%$ to $28.6\%$ ($95\%$ CI on the difference $[11.8, 18.1]$) and raises success for every policy.
On $1{,}465$ on-route states nearest the midpoint of never-deviating episodes, clearing lowers pooled success from $76.0\%$ to $48.3\%$ and lowers success for every policy.

\paragraph{History Length.}
We vary history length by keeping only the $k$ most recent steps,
\begin{equation}
H^{(k)}_t = \{(o_s, a_s)\}_{s=t-k}^{t-1},
\quad k \in \{0, \ldots, t\},
\label{eq:truncation}
\end{equation}
so that $k = 0$ is clearing, written $H^{(\emptyset)}_t$, and $k = t$ is the full-length retained history.
Figure~\ref{fig:history_length} plots each length's success rate minus that with clearing ($\Delta$ success) for the on-route states and the $779$ delayed states, each within its own population.
On route, pooled success rises with $k$, from $9.3$ points above clearing at $k = 2$ to $27.7$ points with the full-length retained history.
After a deviation, every retained length has lower pooled success than clearing, and this gap, which we call the \emph{deficit}, widens from $3.5$ points at $k = 2$ to $13.2$ points with the full-length retained history.

\paragraph{History Content.}
If the off-route frames accumulated after a deviation mislead the policy, removing them should restore recovery.
At the same $779$ delayed states, we therefore retain only the history up to and including the deviation onset, $H_{t_d+1}$ (Figure~\ref{fig:history_content}).
This raises pooled success from $13.1\%$ to $16.3\%$, leaving a deficit of $10.0$ points ($95\%$ CI $[6.5, 13.5]$).
The deficit persists when the history ends before its first frame beyond $0.5$\,m from the route, and an on-route test suggests that the discontinuity between history and current frame does not explain it (Appendix~\ref{app:analysis-details}).
Removing off-route frames recovers only a fraction of the loss: the harm lies mainly outside what they depict.

\subsection{Behavioral Inertia}\label{sec:analysis:inertia}

After a deviation, shuffling the full-length retained history or swapping in one from another scene leaves $13.0$ and $12.5$ of the $13.2$-point deficit, whereas on route both lower success for every policy; length-matched copies of the current frame or noise leave $8.7$ and $7.7$ points (Appendix~\ref{app:analysis-details}).

\begin{figure}[t]
\centering
\captionsetup[sub]{skip=1pt}
\begin{minipage}{0.9\linewidth}
\begin{subfigure}[b]{0.3249\linewidth}
\includegraphics[width=\linewidth]{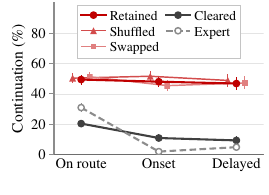}
\caption{Action continuation}\label{fig:continuation}
\end{subfigure}\hfill
\begin{subfigure}[b]{0.3249\linewidth}
\includegraphics[width=\linewidth]{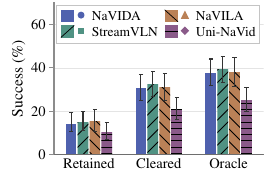}
\caption{Oracle selection}\label{fig:greedy_selection}
\end{subfigure}\hfill
\begin{subfigure}[b]{0.3249\linewidth}
\includegraphics[width=\linewidth]{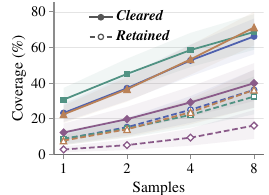}
\caption{Sampling coverage}\label{fig:sampling_coverage}
\end{subfigure}
\end{minipage}
\caption{\textbf{Retained history keeps the policy from turning; after clearing, sampling finds more successes.} (a) After a forward step, four policies pooled. (b) Onset states; oracle: the successful rollout. (c) Onset states whose retained greedy rollout failed. $95\%$ CIs: bootstrap (a,\,c), Wilson (b).}
\label{fig:inertia_recovery}

\end{figure}

\paragraph{Action Continuation.}
We take states right after a forward step and record whether the policy's next action chunk continues without a turn, counting a stop as no turn.
We compare this continuation rate with that of a shortest-path expert queried at the same pose over the same number of steps (Figure~\ref{fig:continuation}).
On route, the expert continues in $31.0\%$ of states, and the policy in $49.5\%$ with retained history and $20.4\%$ after clearing.
After a deviation, the expert almost always turns, continuing in only $1.9\%$ of onset and $4.8\%$ of delayed states, whereas the policy with retained history still continues in $48.1\%$ and $46.8\%$.
After clearing, the continuation rate falls to $10.9\%$ and $9.3\%$, and each policy shows the same pattern.
Shuffled and swapped histories keep the pooled rate close to that of the full-length retained history; per policy they shift it by up to $25$ points in either direction, but for every policy it stays above clearing and the expert (Appendix~\ref{app:analysis-details}).
After a deviation the state calls for a turn, yet the retained history keeps the policy from turning; we call this behavioral inertia.

\subsection{Beyond Clearing}\label{sec:analysis:recovery}

Choosing correctly between retaining and clearing would not rescue most deviations: clearing rescues 195 of the 790 onset states where the retained rollout fails but breaks 58 of the 125 where it succeeds, and even a hindsight oracle leaves $65.0\%$ of states unresolved (Figure~\ref{fig:greedy_selection}).

\paragraph{Sampling Coverage.}
On a separate subset of $320$ onset states where the retained greedy rollout failed, $80$ per policy, we draw eight rollouts per condition (Figure~\ref{fig:sampling_coverage}; Appendix~\ref{app:analysis-details}).
Coverage, written pass@$k$ for $k$ rollouts, is the share of states with at least one successful rollout.
With eight samples, pooled coverage rises from $30.3\%$ with retained history to $61.6\%$ after clearing, yet $38.4\%$ of states still have no successful cleared rollout.
This offline coverage assumes, unlike deployment, that a success can be recognized among the candidates.
Deciding when to clear, and turning successes that appear only under sampling into a single greedy rollout, are the two problems that the deviation judge and the recovery policy of Section~\ref{sec:method} address.

\par
\FloatBarrier

%% file: sec/4.method.tex
\section{Method}\label{sec:method}

\begin{figure}[t]
\centering
\includegraphics[width=\linewidth]{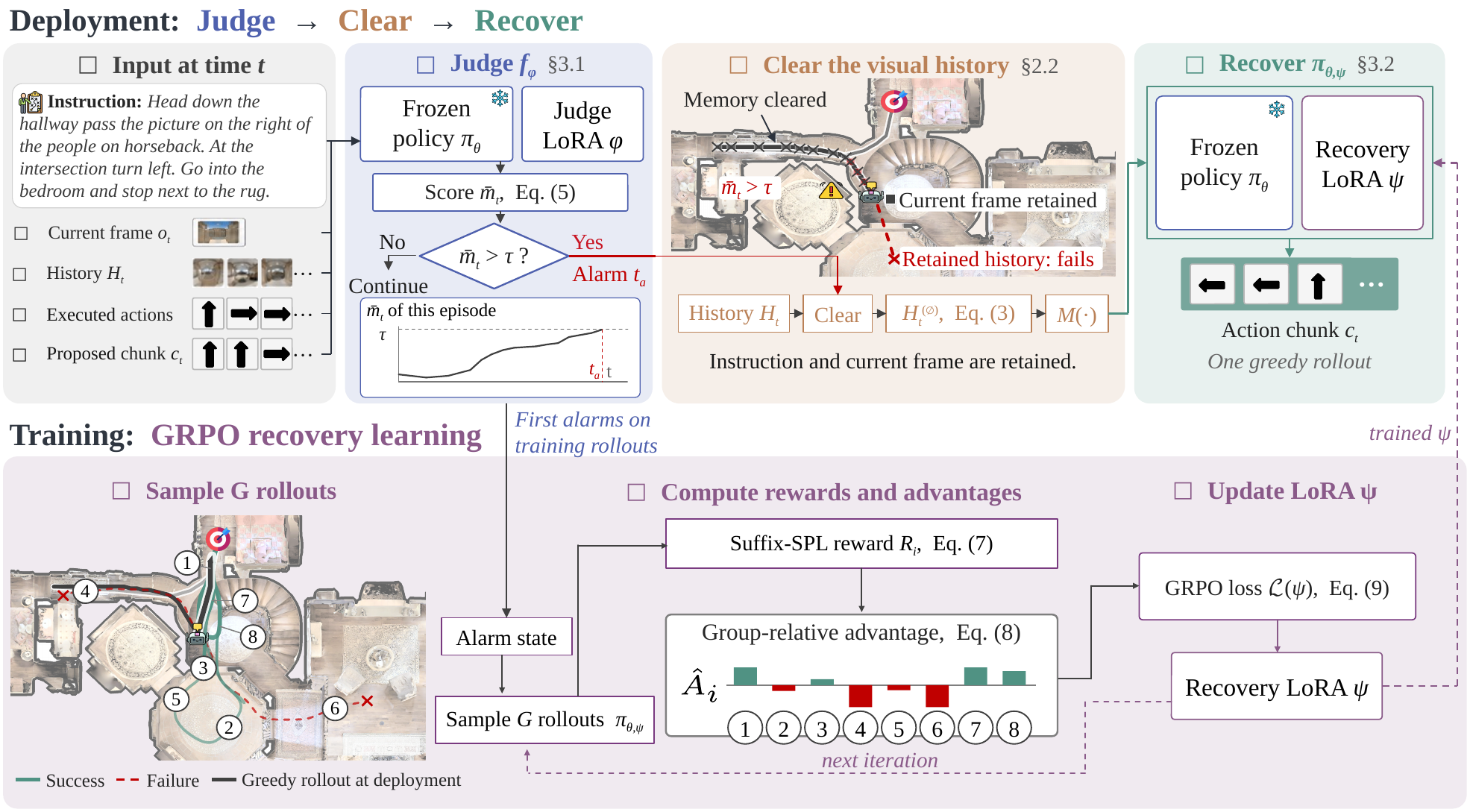}
\caption{\textbf{Overview of \method{}.} Top: at deployment, the deviation judge $f_\phi$ raises an alarm when the deviation score $\bar m_t$ exceeds the conformal threshold $\tau$, clearing empties the visual history, and the recovery policy $\pi_{\theta,\psi}$ acts in one greedy rollout. Bottom: in training, $G$ rollouts are sampled from each alarm state and scored by the suffix-SPL reward, and their group-relative advantages update only the LoRA $\psi$. Scenes and the score inset show one episode of the trained system, while the sampled rollouts are schematic and the advantage bars come from a group resampled after training.}
\label{fig:method}
\end{figure}

\method{} answers the two questions left by Section~\ref{sec:analysis}: when to clear the history (\S\ref{sec:method:judge}), since clearing helps only off route (Section~\ref{sec:analysis:history}), and how to recover after clearing (\S\ref{sec:method:recovery}), since sampling finds successes a greedy rollout misses (Section~\ref{sec:analysis:recovery}).
The deviation judge $f_\phi$ and the recovery policy $\pi_{\theta,\psi}$ add LoRA adapters $\phi$ and $\psi$ to the backbone of the frozen policy $\pi_\theta$~\citep{hu2022lora}.
At the judge's alarm, clearing empties the visual history and the recovery policy takes over (Figure~\ref{fig:method}).

\subsection{Deviation Judge}\label{sec:method:judge}

\paragraph{Prediction Target and Input.}
The route $\mathcal{R}$ is unavailable at deployment, so the judge predicts deviation from the instruction and the agent's own observations and actions.
At each decision point $t$, $\pi_\theta$ proposes the chunk $c_t$ of Section~\ref{sec:analysis:prelim}.
Brief excursions that return to the route occur mostly in episodes that succeed, so a LOST label requires the agent to be off route now and at half or more of a set of later steps $V_t$ within a fixed horizon.
On rollouts of $\pi_\theta$ in the training splits, each decision point receives a label from $h_t$ and $S_d$ of \eqref{eq:deviation}:
\begin{equation}
y_t=
\begin{cases}
\mathrm{LOST}, & \text{if } h_t>S_d \text{ and } \bigl|\{\,u\in V_t : h_u>S_d\,\}\bigr|\ge\tfrac12\,|V_t|,\\
\mathrm{OK}, & \text{otherwise},
\end{cases}
\label{eq:label}
\end{equation}
where $V_t$ is empty when too few steps remain (Appendix~\ref{app:method}).
The judge reads $E_t=(I,o_t,H_t,c_t)$, taking sampled frames and recent actions from $H_t$.
We include the proposed chunk because it shows where the agent is about to go.
It outputs $p_\phi(y\mid E_t)$ for $y\in\{\mathrm{OK},\mathrm{LOST}\}$ and is trained by cross-entropy on these labels.
Stop actions carry no label, and chunks that contain a stop are not judged.

\paragraph{Score and Threshold.}
The deviation score $\bar m_t$ averages the judge's log-odds margin $m_t$ over recent decision points to suppress single-point noise:
\begin{equation}
m_t=\log\frac{p_\phi(\mathrm{LOST}\mid E_t)}{p_\phi(\mathrm{OK}\mid E_t)},
\qquad
\bar m_t=\frac{1}{|\mathcal{W}_t|}\sum_{s\in\mathcal{W}_t} m_s,
\label{eq:score}
\end{equation}
where $\mathcal{W}_t$ holds the last $W$ judged decision points up to $t$.
Clearing on route lowers success (Figure~\ref{fig:history_length}), so we use split conformal prediction~\citep{vovk2005alrw,angelopoulos2021gentle} to bound how often an episode that $\pi_\theta$ would complete raises an alarm.
Each of $n$ held-out calibration episodes that $\pi_\theta$ completes is scored by $Z_e=\max_t\bar m_t$, the closest it comes to an alarm.
With the scores sorted as $Z_{(1)}\le\dots\le Z_{(n)}$ and a level $\alpha$, the threshold and the alarm step are
\begin{equation}
\tau=Z_{(\lceil (n+1)(1-\alpha)\rceil)},
\qquad
t_a=\min\{\,t:\bar m_t>\tau\,\},
\label{eq:conformal}
\end{equation}
where $t$ ranges over judged decision points and $t_a=\infty$ if the score never exceeds $\tau$.
Until $t_a$ the agent is exactly $\pi_\theta$.
A new episode that $\pi_\theta$ completes is therefore interrupted with probability at most $\alpha$ if it is exchangeable with the calibration episodes.
Unlike $t_d$, the alarm step uses only what is available to the agent and can precede $t_d$.

\paragraph{Clearing and Handover.}
An alarm triggers clearing, which reduces the behavioral inertia of Section~\ref{sec:analysis:inertia} (Figure~\ref{fig:continuation}).
At $t_a$, clearing replaces the retained history $H_{t_a}$ with $H^{(\emptyset)}_{t_a}$ of \eqref{eq:truncation} while keeping the instruction and the current frame.
The recovery policy $\pi_{\theta,\psi}$ proposes the next chunk and acts greedily until the episode ends.
Its history re-accumulates from $t_a$, so for $t>t_a$ it reads $H^{(t-t_a)}_t$.
Clearing happens at most once per episode, as in training (\S\ref{sec:method:recovery}).
Algorithm~\ref{alg:deploy} summarizes the loop, and Section~\ref{sec:experiments} compares this timing with other triggers.

\begin{algorithm}[tbp]
\caption{\method{} at deployment: alarm, history clearing and handover to the recovery policy}
\label{alg:deploy}
\renewcommand{\algorithmiccomment}[1]{\hfill$\triangleright$~#1}
\begin{algorithmic}[1]
\REQUIRE instruction $I$, frozen policy $\pi_\theta$, recovery policy $\pi_{\theta,\psi}$ (\S\ref{sec:method:recovery})
\REQUIRE judge $f_\phi$, window $W$, threshold $\tau$ of \eqref{eq:conformal}
\STATE $\pi \leftarrow \pi_\theta$; \ $H \leftarrow \emptyset$
\WHILE{the episode has not ended}
  \STATE $c_t \leftarrow$ greedy decoding of $\pi(\cdot \mid I, o_t, M(H))$
  \IF{$\pi = \pi_\theta$ \AND $c_t$ contains no stop}
    \STATE $E_t \leftarrow (I, o_t, H, c_t)$; \ compute $m_t$ and $\bar m_t$ by \eqref{eq:score}
    \IF{$\bar m_t > \tau$}
      \STATE $H \leftarrow \emptyset$; \ $\pi \leftarrow \pi_{\theta,\psi}$ \COMMENT{first alarm, $t=t_a$}
      \STATE $c_t \leftarrow$ greedy decoding of $\pi_{\theta,\psi}(\cdot \mid I, o_t, M(\emptyset))$
    \ENDIF
  \ENDIF
  \STATE execute $c_t$, appending each step's $(o, a)$ to $H$
\ENDWHILE
\end{algorithmic}
\end{algorithm}

\subsection{Recovery Policy}\label{sec:method:recovery}

\paragraph{Learning from Cleared States.}
After clearing, repeated sampling finds a success at far more states than a single rollout does (Figure~\ref{fig:sampling_coverage}).
This coverage assumes that a success can be recognized among candidates, which deployment cannot do, so the recovery policy must turn these successes into a single greedy rollout.
Reinforcement learning with verifiable rewards can make sampled behaviors more reliable~\citep{yue2025does}, so we train $\pi_{\theta,\psi}$ with GRPO~\citep{shao2024deepseekmath}.
Training starts from the first alarm on rollouts of $\pi_\theta$ in the training splits, successful ones included, with the threshold given in Appendix~\ref{app:method}.
From each alarm state we sample $G$ rollouts $\zeta_1,\dots,\zeta_G$ with $\pi_{\theta,\psi}$, each beginning with the same clearing as at deployment and running without the judge.

\paragraph{Suffix-SPL Reward.}
Only the suffix after the alarm can be credited to the recovery policy, so the reward applies SPL~\citep{anderson2018evaluation} to the suffix alone.
With $d_0$ the geodesic distance from the alarm state to the goal, rollout $\zeta_i$ earns the suffix-SPL reward
\begin{equation}
R_i=\mathbf{1}[\mathrm{succ}_i]\,\Bigl(\lambda+(1-\lambda)\,\frac{d_0}{\max(d_0,\,L_i)}\Bigr),
\label{eq:reward}
\end{equation}
where $\mathrm{succ}_i$ indicates that $\zeta_i$ stops within the success radius and $L_i$ is its path length.
Shorter successes earn more when $L_i>d_0$, and the floor $\lambda\in(0,1)$ keeps every success above every failure.
GRPO normalizes rewards within the group, which gives a baseline without a learned critic:
\begin{equation}
\hat A_i=\frac{R_i-\operatorname{mean}(R_1,\dots,R_G)}{\operatorname{std}(R_1,\dots,R_G)+\epsilon},\qquad \epsilon=10^{-6}.
\label{eq:advantage}
\end{equation}
We update only the recovery adapter $\psi$ by minimizing the loss $\mathcal{L}(\psi)$:
\begin{equation}
\mathcal{L}(\psi)=\frac{1}{N}\sum_{i,j} w_{i,j}
\Bigl[\ell\bigl(c_{i,j},\hat A_i\bigr)+\beta\,\hat D_{i,j}\Bigr],
\label{eq:grpo}
\end{equation}
where $c_{i,j}$ is the $j$-th chunk of $\zeta_i$, $\ell$ is the clipped surrogate loss that GRPO inherits from PPO~\citep{shao2024deepseekmath,schulman2017proximal},
and $\hat D_{i,j}$ estimates $\KL(\pi_{\theta,\psi}\,\|\,\pi_\theta)$, with weights and normalization as in the appendix.
Groups of equal rewards have zero advantage and are discarded.
In practice almost all of these groups contain only failures, and clearing leaves fewer of them than retaining the history.
Section~\ref{sec:experiments} measures how clearing affects training.

%% file: sec/5.experiments.tex
\section{Experiments}\label{sec:experiments}

\subsection{Experimental Setup}\label{sec:exp:setup}
\paragraph{Implementation Details.}
We evaluate on the full val-unseen splits of R2R-CE ($1{,}839$ episodes) and RxR-CE ($3{,}669$ episodes) with NE, OS, SR, SPL and, on RxR-CE, nDTW.
The frozen policy, NaVIDA-Qwen3VL, follows the official training recipe of NaVIDA~\citep{zhu2026navida} and only replaces its backbone with Qwen3-VL-4B-Instruct.
It is not among the four policies of Section~\ref{sec:analysis}, so Section~\ref{sec:exp:ablation} repeats the off-route clearing test on it.
The judge and the recovery policy are trained on the training splits, the latter with GRPO for 20 iterations from alarm states.
The threshold is calibrated per dataset at $\alpha=10\%$ for R2R-CE and $7.5\%$ for RxR-CE, the levels with the highest closed-loop SR of those evaluated on val-unseen.
All policies decode greedily, and paired comparisons use the exact McNemar test on episode outcomes; Appendices~\ref{app:method} and~\ref{app:alarm-level} give the details.

\subsection{Main Results}\label{sec:exp:main}

\begin{table}[t]
{%
\centering
\caption{Comparison with prior methods on the R2R-CE and RxR-CE val-unseen splits. \textbf{Bold} and \underline{underline} mark the best and second-best results in each column. Pano., Odo., Depth and S.RGB denote panoramic view, odometry, depth and single RGB input. $^\dagger$Recovery-oriented method.}
\label{tab:main}
\footnotesize
\setlength{\tabcolsep}{1.8pt}\setlength{\aboverulesep}{0pt}\setlength{\belowrulesep}{0pt}\setlength{\extrarowheight}{0.1ex}
\resizebox{\linewidth}{!}{%
\begin{tabular}{lc<{\hspace{2pt}}|>{\hspace{1.5pt}}cccc<{\hspace{1.5pt}}|>{\hspace{1.5pt}}cccc<{\hspace{1.5pt}}|>{\hspace{1.5pt}}cccc}
\toprule
\multirow{2}{*}{Method} & \multirow{2}{*}{Params} & \multicolumn{4}{c|}{Observation} & \multicolumn{4}{c|}{R2R-CE Val-Unseen} & \multicolumn{4}{c}{RxR-CE Val-Unseen}\\
\cmidrule(lr){3-6} \cmidrule(lr){7-10} \cmidrule(lr){11-14}
 & & {\scriptsize Pano.} & {\scriptsize Odo.} & {\scriptsize Depth} & {\scriptsize S.RGB} & NE$\downarrow$ & OS$\uparrow$ & SR$\uparrow$ & SPL$\uparrow$ & NE$\downarrow$ & SR$\uparrow$ & SPL$\uparrow$ & nDTW$\uparrow$\\
\midrule
{\color{dark-gray}\textit{Panoramic or depth input}} & & & & & & & & & & & & & \\
HPN+DN~\pub{ICCV21}~\citep{krantz2021waypoint} & -- & \cm & \cm & \cm &  & 6.31 & 40.0 & 36.0 & 34.0 & -- & -- & -- & -- \\
CMA~\pub{CVPR22}~\citep{hong2022bridging} & -- & \cm & \cm & \cm &  & 6.20 & 52.0 & 41.0 & 36.0 & 8.76 & 26.6 & 22.2 & 47.1 \\
ETPNav~\pub{TPAMI24}~\citep{an2024etpnav} & -- & \cm & \cm & \cm &  & 4.71 & 65.0 & 57.0 & 49.0 & 5.64 & 54.8 & 44.9 & 61.9 \\
NavMorph~\pub{ICCV25}~\citep{yao2025navmorph} & -- &  &  & \cm & \cm & 5.75 & 56.9 & 47.9 & 33.2 & 8.85 & 30.8 & 22.8 & 44.2 \\
SC$^2$-WM$^\dagger$~\pub{ICML26}~\citep{yao2026sc2wm} & -- & \cm & \cm & \cm &  & 4.25 & 70.0 & 66.0 & 54.0 & 5.72 & 60.0 & 50.0 & 67.0 \\
\midrule
{\color{dark-gray}\textit{Monocular RGB, recovery-oriented}} & & & & & & & & & & & & & \\
CorrectNav~\pub{AAAI26}~\citep{yu2025correctnav} & 7B &  &  &  & \cm & 4.24 & 67.5 & 65.1 & \textbf{62.3} & 4.09 & 69.3 & \underline{63.3} & \textbf{75.2} \\
BudVLN~\pub{arXiv26}~\citep{budvln2026} & 7B &  &  &  & \cm & 4.74 & 65.6 & 57.6 & 51.1 & 5.79 & 56.1 & 46.6 & 63.2 \\
SACA~\pub{arXiv26}~\citep{li2026saca} & 8B &  &  &  & \cm & 4.19 & 69.3 & 64.7 & 56.9 & 4.75 & 62.1 & 51.7 & 66.0 \\
DecoVLN~\pub{CVPR26}~\citep{xin2026decovln} & 7B &  &  &  & \cm & 5.01 & 63.5 & 56.3 & 50.5 & 5.73 & 54.2 & 46.3 & 63.5 \\
ReflectVLN~\pub{arXiv26}~\citep{wang2026reflectvln} & 6B &  &  &  & \cm & 4.19 & 67.3 & 62.8 & 58.5 & \underline{3.98} & 66.0 & 57.2 & -- \\
\midrule
{\color{dark-gray}\textit{Monocular RGB, general-purpose}} & & & & & & & & & & & & & \\
NaVid~\pub{RSS24}~\citep{zhang2024navid} & 7B &  &  &  & \cm & 5.47 & 49.1 & 37.4 & 35.9 & -- & -- & -- & -- \\
Uni-NaVid~\pub{RSS25}~\citep{zhang2024uninavid} & 7B &  &  &  & \cm & 5.58 & 53.3 & 47.0 & 42.7 & 6.24 & 48.7 & 40.9 & -- \\
NaVILA~\pub{RSS25}~\citep{cheng2024navila} & 8B &  &  &  & \cm & 5.22 & 62.5 & 54.0 & 49.0 & 6.77 & 49.3 & 44.0 & 58.8 \\
StreamVLN~\pub{ICRA26}~\citep{wei2025streamvln} & 7B &  &  &  & \cm & 4.90 & 63.6 & 56.4 & 50.2 & 5.65 & 54.4 & 45.4 & 63.7 \\
NavFoM~\pub{ICLR26}~\citep{zhang2026navfom} & 7B &  &  &  & \cm & 5.01 & 64.9 & 56.2 & 51.2 & 5.51 & 57.4 & 49.4 & 60.2 \\
JanusVLN~\pub{ICLR26}~\citep{zeng2026janusvln} & 7B &  &  &  & \cm & 4.78 & 65.2 & 60.5 & 56.8 & 6.06 & 56.2 & 47.5 & 62.1 \\
DualVLN~\pub{ICLR26}~\citep{wei2026dualvln} & 7B &  &  &  & \cm & \underline{4.05} & 70.7 & 64.3 & 58.5 & 4.58 & 61.4 & 51.8 & 70.0 \\
FutureNav-4B~\pub{arXiv26}~\citep{futurenav} & 4B &  &  &  & \cm & 4.24 & 70.2 & 65.4 & \underline{61.3} & 4.60 & 60.9 & 52.8 & 68.0 \\
FutureNav-8B~\pub{arXiv26}~\citep{futurenav} & 8B &  &  &  & \cm & 4.29 & 69.2 & 64.3 & 60.3 & 4.26 & 63.9 & 54.8 & 69.1 \\
Qwen-RobotNav-4B~\pub{arXiv26}~\citep{qwenrobotnav} & 4B &  &  &  & \cm & 4.22 & \underline{73.6} & \underline{66.9} & 60.5 & 4.15 & \underline{71.3} & 61.5 & 68.6 \\
Qwen-RobotNav-8B~\pub{arXiv26}~\citep{qwenrobotnav} & 8B &  &  &  & \cm & 4.36 & 72.7 & 65.7 & 59.6 & 4.16 & \textbf{73.4} & \textbf{63.5} & 69.9 \\
\midrule
\rowcolor{blue!10}
\method{} (Ours) & 4B &  &  &  & \cm & \textbf{3.56} & \textbf{75.9} & \textbf{69.0} & \textbf{62.3} & \textbf{3.75} & 70.9 & 60.8 & \underline{72.4} \\
\bottomrule
\end{tabular}}
}
\end{table}

\paragraph{Comparison with Prior Methods.}
On R2R-CE (Table~\ref{tab:main}), \method{} is best in every column, with an SR of 69.0 against 66.9 for the next method, Qwen-RobotNav-4B, and an SPL tying CorrectNav at 62.3.
On RxR-CE it has the lowest NE and second-highest nDTW, with SR and SPL within 2.7 points of the best, Qwen-RobotNav-8B.
Over the frozen base, the adapters raise SR from 63.8 to 69.0 on R2R-CE and from 64.9 to 70.9 on RxR-CE, and SPL from 59.0 to 62.3 and 55.9 to 60.8.

\begin{figure}[t]
\centering
\begin{minipage}[t]{0.47\textwidth}
\centering
\vspace{0pt}
\includegraphics[width=0.68\linewidth]{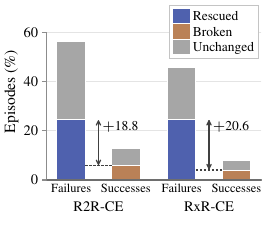}

\captionof{figure}{\textbf{Judge alarms and their outcomes.} Bars: share of base failures or successes with an alarm; colored: share rescued or broken; arrows: their difference, not an SR change.}
\label{fig:judge_outcomes}
\end{minipage}\hfill
\begin{minipage}[t]{0.47\textwidth}
\centering
\vspace{0pt}
\includegraphics[width=0.68\linewidth]{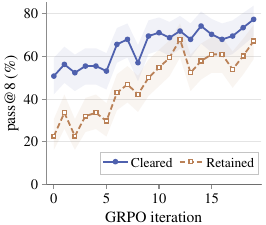}

\captionof{figure}{\textbf{Training with cleared and retained history.} pass@8 on the $128$ alarm states of each GRPO iteration, the same states for both runs. Shaded: $95\%$ Wilson intervals.}
\label{fig:training_efficiency}
\end{minipage}

\end{figure}

\paragraph{Judge Alarms.}
Figure~\ref{fig:judge_outcomes} shows where the judge alarms and what the alarm changes.
The judge alarms in 56.3\% of the episodes the base fails on R2R-CE and 45.8\% on RxR-CE, but in only 13.0\% and 8.0\% of those it completes.
Over all episodes, the full system rescues 24.6\% of the base failures and breaks 5.8\% of the base successes on R2R-CE (164 against 68 episodes, exact McNemar test, $p=2\times10^{-10}$), and 24.7\% against 4.1\% on RxR-CE (317 against 97, $p=2\times10^{-28}$).
Appendix~\ref{app:judge-counts} gives the counts, the recovery policy's effect on alarmed episodes and the alarm timing.

\paragraph{Zero-Shot Object Navigation.}
Zero-shot on HM3D-OVON val-unseen, with both adapters and the threshold unchanged, \method{} raises the base's SR from 40.7 to 57.2 and SPL from 21.7 to 35.4, the best among the compared methods, trained on HM3D-OVON or not (Table~\ref{tab:ovon}, Appendix~\ref{app:results}).

\begin{figure}[t]
\centering
\includegraphics[width=\linewidth,height=0.82\textheight,keepaspectratio]{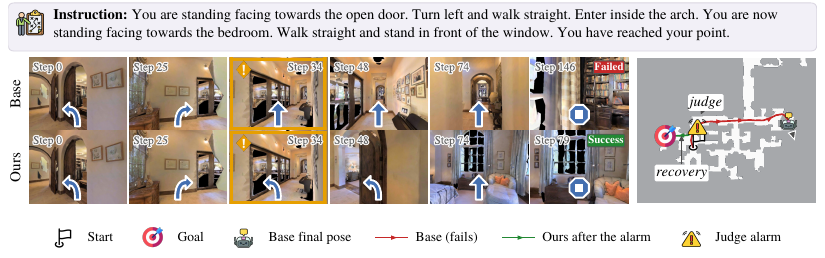}
\caption{\textbf{Qualitative result on RxR-CE.} Rows share frames until the alarm (amber border), after which \method{} clears the history and the recovery policy acts; last column: end of each run.}
\label{fig:qualitative}

\end{figure}

\paragraph{Qualitative Results.}
Figure~\ref{fig:qualitative} shows an RxR-CE episode with an alarm; Appendix~\ref{app:qualitative} shows more.

\subsection{Ablation Study}\label{sec:exp:ablation}

\begin{table}[t]
\centering
\begin{minipage}[t]{0.32\textwidth}
\centering
\caption{Core components.}
\label{tab:ablation-components}
\footnotesize
\setlength{\tabcolsep}{1.5pt}\setlength{\aboverulesep}{0pt}\setlength{\belowrulesep}{0pt}\setlength{\extrarowheight}{0.1ex}
\begin{tabular}{ccccc}
\toprule
Judge & Clear & Recovery & SR$\uparrow$ & SPL$\uparrow$ \\
\midrule
\xm & \xm & \xm & 63.8 & 59.0 \\
\xm & \xm & \cm & 59.9 & 54.5 \\
\cm & \cm & \xm & 65.6 & 59.0 \\
\cm & \xm & \cm & 67.4 & 60.5 \\
\rowcolor{blue!10}
\cm & \cm & \cm & 69.0 & 62.3 \\
\bottomrule
\end{tabular}
\end{minipage}\hfill
\begin{minipage}[t]{0.32\textwidth}
\centering
\caption{Alarm trigger, SR.}
\label{tab:ablation-trigger}
\footnotesize
\setlength{\tabcolsep}{2.5pt}\setlength{\aboverulesep}{0pt}\setlength{\belowrulesep}{0pt}\setlength{\extrarowheight}{0.1ex}
\begin{tabular}{lcc}
\toprule
 & Clear & \\
Trigger & only & + Recovery \\
\midrule
Random & 58.2 & 60.4 \\
\rowcolor{blue!10}
Judge & 65.6 & 69.0 \\
Ground truth & 69.0 & 71.2 \\
\bottomrule
\end{tabular}
\end{minipage}\hfill
\begin{minipage}[t]{0.32\textwidth}
\centering
\caption{Recovery training.}
\label{tab:ablation-training}
\footnotesize
\setlength{\tabcolsep}{2.5pt}\setlength{\aboverulesep}{0pt}\setlength{\belowrulesep}{0pt}\setlength{\extrarowheight}{0.1ex}
\begin{tabular}{lcc}
\toprule
Training & SR$\uparrow$ & SPL$\uparrow$ \\
\midrule
None (clear only) & 65.6 & 59.0 \\
SFT & 62.9 & 57.6 \\
GRPO$_{\mathrm{start}}$ & 66.2 & 59.5 \\
GRPO$_{\mathrm{keep}}$ & 66.2 & 60.3 \\
\rowcolor{blue!10}
GRPO (ours) & 69.0 & 62.3 \\
\bottomrule
\end{tabular}
\end{minipage}
\end{table}

\paragraph{Components.}
Tables~\ref{tab:ablation-components}--\ref{tab:ablation-training} use R2R-CE, and Appendix~\ref{app:results} varies the alarm level and the judge query interval.
In Table~\ref{tab:ablation-components}, without the judge, the recovery policy acts from the first step and SR falls to 59.9, below the base ($p=3\times10^{-12}$ against the full system).
Without the recovery policy, clearing at the alarm still raises SR above the base, to 65.6 ($p=0.018$), but below the full system ($p=1\times10^{-4}$): most of the gain comes from the recovery policy.
Without clearing, the recovery policy takes over the retained history and reaches 67.4, not significantly below the full system ($p=0.24$): a policy trained to recover copes with the retained history better than the untrained policies of Section~\ref{sec:analysis}, and clearing still matters in training (below).

\paragraph{Triggers.}
Table~\ref{tab:ablation-trigger} changes when clearing starts, with and without the recovery policy.
A random trigger with as many alarms at the same relative depths (Appendix~\ref{app:judge-counts}) lowers SR below the base, to 58.2 with clearing alone and 60.4 with the recovery policy.
A ground-truth trigger at the first step with $h_t>S_d$ applies the intervention of Section~\ref{sec:analysis:history} to the deployed base: clearing alone adds 5.2 points to SR, as for the policies of Section~\ref{sec:analysis}.
With the recovery policy it reaches 71.2, as it catches 76.3\% of base failures against 56.3\% for the judge; the headroom lies in detection.

\paragraph{Recovery Training.}
After clearing, sampling finds more recoveries than a greedy rollout, as in Section~\ref{sec:analysis:recovery}: on the 527 alarmed R2R-CE episodes, at least one of eight samples succeeds in 56.0\% against 35.1\% for the greedy rollout, and GRPO raises greedy success to 47.1\%.
In Table~\ref{tab:ablation-training}, SFT on shortest-path demonstrations falls below the base (62.9), as the expert heads for a goal the cleared policy cannot see, while GRPO$_{\mathrm{start}}$ (rollouts from the episode start) and GRPO$_{\mathrm{keep}}$ (history retained in training and at deployment) both give 66.2.
Clearing also makes training more efficient (Figure~\ref{fig:training_efficiency}): at the first iteration, pass@8 is 50.8\% with cleared rollouts against 22.7\% with retained ones, so fewer groups carry no signal, although the two runs end at similar closed-loop SR (Table~\ref{tab:ablation-train-full}).

\begin{wraptable}[4]{r}{0.35\textwidth}

\centering
\caption{Parameters and latency}
\label{tab:efficiency}

\scriptsize
\setlength{\tabcolsep}{2.2pt}\setlength{\aboverulesep}{0pt}\setlength{\belowrulesep}{0pt}\setlength{\extrarowheight}{0.1ex}
\begin{tabular}{lcc}
\toprule
Component & Params & Latency \\
\midrule
Base (frozen) & 4.44B & 498\,ms \\
+ Judge & +23.6M & -- \\
+ Recovery & +12.4M & -- \\
\midrule
\rowcolor{blue!10}
Full system & +36.0M {\color{costup}$\uparrow$0.81\%} & 507\,ms {\color{costup}$\uparrow$1.8\%} \\
\bottomrule
\end{tabular}

\end{wraptable}
\paragraph{Efficiency.}
The two adapters increase the parameters of the frozen base by 0.81\% and its time per action by 1.8\% (Table~\ref{tab:efficiency}).
The judge is trained on 200{,}529 decision points and the recovery policy on 20{,}480 rollouts from 2{,}560 alarm states, against 4.92M samples for the base.

%% file: sec/6.conclusion.tex
\section{Conclusion}
We asked how visual history affects recovery after a deviation.
Paired interventions on four general-purpose VLN policies show that the retained history helps on route but hinders recovery off route, where it keeps the policy from turning.
After clearing, sampling often finds successes that a single greedy rollout misses.
\method{} adds two lightweight modules to a frozen policy: a deviation judge decides when to clear the history, and a GRPO-trained recovery policy turns more of these successes into a single greedy rollout.
With $0.81\%$ additional parameters, it raises the success rate of its base by $5.2$ points on R2R-CE and $6.0$ on RxR-CE, and improves zero-shot object-goal navigation without retraining.
The appendix further reports the analysis protocol with per-policy results and history substitutions, ablations of the alarm level, the judge query interval and the training recipe, zero-shot object-goal navigation, and the prompts.
Beyond navigation, many embodied policies act on a history of their own observations and actions; future work could plug \method{} into newer open-source policies, detect deviations earlier, and teach policies to forget from within rather than by clearing.

%% file: sec/8.limitations.tex
\section{Limitations}\label{sec:limitations}
\paragraph{One deployed policy.}
\method{} is deployed on one frozen policy, NaVIDA-Qwen3VL, while the analysis of Section~\ref{sec:analysis} covers four policies with three kinds of visual memory.
Because \method{} keeps the policy frozen and trains only two lightweight LoRA adapters on its backbone, it can be plugged into newer open-source policies that condition on their visual history as they are released, and evaluating it on such policies is a natural next step.

\paragraph{Detection bounds the gain.}
The judge alarms in $56.3\%$ of the episodes the base fails on R2R-CE, and a ground-truth trigger with the same recovery policy reaches an SR of $71.2$ against $69.0$ (Table~\ref{tab:ablation-trigger}).
Alarms also come a median of $13$ atomic steps after the deviation onset on R2R-CE and $21$ on RxR-CE (Appendix~\ref{app:judge-counts}), so earlier and more accurate detection could recover more episodes.

\paragraph{Clearing has a cost.}
Clearing breaks $5.8\%$ of the episodes the base completes on R2R-CE and $4.1\%$ on RxR-CE, and each episode is cleared at most once.

%% file: sec/7.appendix.tex
\clearpage
\section*{Appendix Contents}
\startcontents[appendix]
\begin{tcolorbox}[breakable, colback=black!3, colframe=black!18, boxrule=0.4pt, arc=2mm, left=5mm, right=5mm, top=3mm, bottom=3mm]
\printcontents[appendix]{l}{1}{\setcounter{tocdepth}{2}}
\end{tcolorbox}
\clearpage
\section{Related Work}\label{app:related}

\subsection{Vision-and-Language Navigation and Visual Memory}
Vision-and-language navigation has progressed from path selection on predefined navigation graphs to action execution in continuous environments \citep{anderson2018vln,krantz2020vlnce}.
History-aware models such as HAMT incorporate past observations into cross-modal reasoning \citep{chen2021hamt}.
More recent approaches, including NaVid and NaVILA, build on pretrained vision-language models to predict navigation actions from instructions, current observations, and visual history \citep{zhang2024navid,cheng2024navila,zhu2026navida}.
To keep this history compact, they sample frames, merge visual tokens online, or stream caches \citep{cheng2024navila,zhu2026navida,zhang2024uninavid,wei2025streamvln}: NaVILA and NaVIDA sample a fixed number of past frames uniformly, Uni-NaVid keeps fewer tokens for older frames, and StreamVLN pairs a key--value cache of recent steps with a slowly updated memory.
Other methods select history frames by semantic relevance, visual diversity, and temporal coverage \citep{xin2026decovln}, decouple semantics and spatiality with a dual implicit memory \citep{zeng2026janusvln}, suppress history noise, or compress context \citep{memoryadaptive2024,astranavmemory2026}.
However, memory ablations in several of these works retrain the policy \citep{zhang2024uninavid,wei2025streamvln}, and evaluation largely relies on full-episode navigation performance, without distinguishing the benefits and costs of history between on-route and off-route states.
We instead keep the policies fixed, intervene only on the history they receive, and compare on-route states with states after a deviation.

\subsection{Deviation Detection and Recovery}
Progress monitors estimate how far an agent has advanced toward the goal \citep{ma2019selfmonitoring}, and backtracking methods use such estimates, global signals, or history-aware planning to revisit earlier locations and reconsider the route \citep{ma2019regretful,ke2019tactical,smartway2025}.
The closest prior detector reads path deviation from attention heads of the frozen NaVILA checkpoint of Section~\ref{sec:analysis} and hands control to a separate controller, leaving the VLA context untouched \citep{jeong2026navheads}.
At an alarm, this controller drives the robot back to the last pose judged normal.
Its authors list this unsynchronized context as a limitation.
In ReflectVLN, an execution agent signals a deviation, and a separate intention agent reflects on it to produce a corrective subtask \citep{wang2026reflectvln}.
\method{} also raises its alarm on a frozen policy, using a trained adapter that also reads the chunk the policy proposes next, but it then leaves the agent at its current pose, clears the policy's visual history, and hands control to a recovery adapter on the same backbone.

\subsection{Corrective Training and Reinforcement Learning}
CorrectNav finds where its trained model still deviates on training trajectories and retrains on synthesized corrections, keeping earlier observations as history while supervising only the subsequent corrective trajectory \citep{yu2025correctnav}.
BudVLN identifies conflicts between corrective actions from error states and the original instruction, and returns to valid historical states to construct aligned supervision \citep{budvln2026}.
DecoVLN builds corrective supervision for state--action pairs from geodesic deviation, keeping only samples within a trust region around the expert trajectory \citep{xin2026decovln}, and SACA resamples a repair from the divergence point of a failed trajectory \citep{li2026saca}.
Reinforcement learning with task rewards also improves navigation robustness \citep{navgrpo2026,qi2025vlnr1}, and BudVLN and SACA combine corrective data with GRPO.
These evaluations, however, do not isolate the effect of visual history on recovery under a fixed policy, so improved recovery does not show whether the history supplies useful cues for returning to the instructed route or interferes with adaptation to the current state.
In manipulation, RePO-VLA clears the history only in recovery training data, so that corrective actions condition on the current state rather than on the earlier failure, with no reset or failure detector at deployment \citep{liufu2026repovla}.
\method{} instead keeps the base policy frozen, clears the history at deployment when its judge raises an alarm, and trains a recovery adapter with GRPO from the cleared alarm states.

\subsection{History Dependence in Imitation Learning}
Imitation policies conditioned on past observations and actions can rely on correlates of the expert's past behavior instead of the current observation \citep{dehaan2019causal,wen2020fighting}.
Access to more information can then worsen performance under distribution shift \citep{dehaan2019causal}, and copycat policies predict the expert's previous action instead of the next one \citep{wen2020fighting}.
\citet{codevilla2019exploring} describe an inertia problem in end-to-end driving, where a policy conditioned on its own speed learns that a stopped car tends to stay stopped.
Once a policy leaves the expert distribution, it meets histories that its training data did not cover, and its errors compound \citep{ross2011reduction,sobolmark2026bpp}.
Proposed remedies either change training, through targeted interventions \citep{dehaan2019causal}, removing information about the previous action \citep{wen2020fighting}, and auxiliary speed prediction \citep{codevilla2019exploring}, or replace the raw history with key frames in both training and deployment \citep{sobolmark2026bpp}.
The inertia we study arises from the retained visual history of navigation policies and becomes harmful after a deviation, when the state calls for a change of course.
Most of this harm remains when the history is replaced by length-matched copies of the current frame or by noise (Section~\ref{sec:analysis:inertia}).
Where these remedies reshape training or replace the history throughout an episode, \method{} keeps the base policy frozen and clears its history only after an alarm.

\subsection{Calibrated Thresholds and Reinforcement Fine-Tuning}
Split conformal prediction sets a threshold from held-out calibration scores and bounds the error rate on new data that are exchangeable with them \citep{vovk2005alrw,angelopoulos2021gentle}.
\citet{jeong2026navheads} instead choose attention heads on labeled training episodes and set their threshold by grid search under an empirical false-alarm constraint.
GRPO normalizes rewards within a group of sampled outputs, which gives a baseline without a learned critic \citep{shao2024deepseekmath}, and reinforcement learning with verifiable rewards is effective at making behaviors that a policy can already sample more reliable \citep{yue2025does}.
\method{} calibrates its threshold on episodes that the frozen policy completes, bounding how often such episodes raise an alarm, and trains its LoRA recovery adapter \citep{hu2022lora} with GRPO only from cleared alarm states, rewarding the part of each rollout after the alarm.

\FloatBarrier
\section{Details of the Motivating Analyses}\label{app:analysis-details}

\subsection{State Populations and Uncertainty}
All interventions keep the navigation policy fixed and retain the current observation.
Each episode is capped at $400$ steps.
NaVIDA samples eight frames uniformly from $H_t$ and NaVILA seven, each alongside the current frame; StreamVLN keeps a key--value cache of its most recent $32$ steps, generated actions included, and an eight-frame block for each earlier window; and Uni-NaVid merges visual tokens online.
For NaVIDA and NaVILA, increasing $k$ beyond eight spreads a fixed number of subsampled frames over a longer time span.
Decoding is greedy except in the sampling study of Section~\ref{sec:analysis:recovery}, and the simulator is deterministic, so the rollout that retains the history is the remainder of the original trajectory.
NaVIDA, StreamVLN and NaVILA run from the \texttt{NaVIDA-3B}, \texttt{StreamVLN\_v1\_3} and \texttt{navila-llama3-8b-8f} checkpoints.
Uni-NaVid runs from \texttt{uninavid-7b-interim-from-dexbotic}, an interim 7B checkpoint from Dexbotic. We did not use the official Uni-NaVid weights.
Deviation onset uses the route-deviation threshold $S_d = 2$\,m of Section~\ref{sec:analysis:prelim}.
The delayed condition uses the first available decision point at least $10$ atomic steps after onset.
At $749$ of the $779$ delayed states ($96.1\%$), the agent is still more than $2$\,m from the route.
On-route controls are sampled from episodes that never deviate, at the decision point nearest the midpoint of each episode.
Across the four policies, $1{,}465$ of the $2{,}400$ rollouts never deviate; of the $935$ that deviate, $20$ do so after their last decision point, which leaves $915$ onset states.
Table~\ref{tab:analysis-populations} records the distinct populations used in the figures.
State-paired success differences use $2{,}000$ bootstrap resamples with seed $41$.
Within a policy, each of the on-route, onset and delayed populations holds at most one state per episode, so the state bootstrap is also an episode bootstrap.
Pooled intervals resample the states of the four policies independently; resampling episodes jointly across the policies moves the bounds of the pooled success differences by less than one point.
The $95\%$ intervals are pointwise, not simultaneous bands.
Equivalence comparisons use $90\%$ paired bootstrap intervals and margins of $\pm8$ pp per policy and $\pm5$ pp pooled.
We set these margins after reading the estimates of the shuffled and swapped success comparisons on the same $779$ delayed states.
An interval contained within the corresponding margin supports equivalence at that resolution, rather than exact equality.

\begin{table}[!htbp]
\centering
\caption{State counts for the analysis panels. After-deviation action observations include both onset and delayed states from some of the same episodes.}
\label{tab:analysis-populations}
\small
\setlength{\tabcolsep}{4pt}
\begin{tabular}{lrrrr}
\toprule
Population & NaVIDA & StreamVLN & NaVILA & Uni-NaVid\\
\midrule
On-route history controls & 359 & 337 & 402 & 367\\
Delayed history interventions & 205 & 210 & 163 & 201\\
Greedy onset interventions & 238 & 248 & 198 & 231\\
Sampling: retained failures & 80 & 80 & 80 & 80\\
Action continuation, onset and delayed & 279 & 302 & 292 & 391\\
\bottomrule
\end{tabular}
\end{table}
\FloatBarrier

\subsection{History Constructions and Comparators}
In Figure~\ref{fig:history_length}, $k$ counts past observations in addition to the current frame, with positions proportional to $\log_2(k+1)$.
All denotes each policy's original history mechanism, including cached context where applicable, and is separated from numeric frame counts by an axis break.
The full-history control in Table~\ref{tab:history-constructions} replays the recorded frames for Uni-NaVid.
For StreamVLN it also replays the frames and forces the recorded answers into the key--value cache at each recorded decision step.
NaVIDA and NaVILA use the standard full-history condition.
Shuffled and swapped paired contrasts use the native full-history reference.
Figure~\ref{fig:history_content} compares the full-history control with prefix + current, both relative to current-frame-only.
The full-length retained history of the main text is this control in Figure~\ref{fig:history_content} and Section~\ref{sec:analysis:history}, and the native reference in the shuffled and swapped contrasts; the two need not give identical estimates.
Prefix + current retains the prefix up to deviation onset and the current observation, removing intervening post-onset history; it is the history $H_{t_d+1}$ of Section~\ref{sec:analysis:history}, which ends with the onset frame.
The 1.5, 1.0 and 0.5\,m prefixes stop before the first frame farther than that distance from the route.
They leave pooled deficits of $9.9$, $8.2$ and $7.4$ points, and the deficit of the onset prefix is significant for three of the four policies (NaVILA: $4.3$, $[-3.7, 12.3]$).
On route, histories whose last frame lies $11$ or $21$ steps before the current frame raise success over current-frame-only by $27.4$, $26.4$, $23.5$ and $8.8$ points, and by $17.6$, $10.1$, $22.5$ and $13.7$ points (NaVIDA, StreamVLN, NaVILA, Uni-NaVid), with every $95\%$ interval excluding zero.
A swapped history is drawn from an on-route episode in another scene, while token-matched noise replaces visual content at matched token count.
Expert + tail replays the dataset expert's actions along the reference route up to the waypoint nearest the current pose, then follows a shortest-path tail to the same pose.
In $359$ of the $779$ delayed states that waypoint is the episode start, so the reconstruction contains no route segment.
Because the tail must reach the current pose, which lies a median $3.85$\,m from the route, its final steps usually lie outside the $S_d$ band.
Expert + tail is therefore not a history free of off-route content, and the main text does not use it as one.
Route-only history keeps only the route segment and is defined on the remaining $420$ states.
The own-history control matches the expert reconstruction's length.
Table~\ref{tab:history-constructions} includes all seventeen tested conditions, including comparisons whose intervals contain zero.

\begin{table}[!htbp]
\centering
\caption{Success-rate differences from current-frame-only, in points; a negative value is a deficit of the same size in the main text. Each cell gives the estimate and its pointwise $95\%$ paired state-bootstrap interval. The default population is $779$ delayed states. Route-only uses $420$ eligible states. Pooled estimates weight states equally.}
\label{tab:history-constructions}
\small
\setlength{\tabcolsep}{3pt}
\renewcommand{\arraystretch}{1.15}
\begin{tabular}{lrrrrr}
\toprule
Construction & NaVIDA & StreamVLN & NaVILA & Uni-NaVid & Pooled\\
\midrule
Prefix 0.5 m & \shortstack[r]{$-7.3$\\$[-14.2, 0.0]$} & \shortstack[r]{$-14.8$\\$[-22.4, -7.6]$} & \shortstack[r]{$-0.6$\\$[-8.6, 7.4]$} & \shortstack[r]{$-5.5$\\$[-11.4, 0.5]$} & \shortstack[r]{$-7.4$\\$[-10.8, -3.7]$}\\[2pt]
Prefix 1.0 m & \shortstack[r]{$-10.7$\\$[-18.5, -3.4]$} & \shortstack[r]{$-12.4$\\$[-19.5, -5.7]$} & \shortstack[r]{$-3.1$\\$[-11.0, 4.9]$} & \shortstack[r]{$-5.5$\\$[-11.4, 0.0]$} & \shortstack[r]{$-8.2$\\$[-11.8, -4.6]$}\\[2pt]
Prefix 1.5 m & \shortstack[r]{$-10.7$\\$[-18.1, -3.4]$} & \shortstack[r]{$-12.4$\\$[-19.5, -5.2]$} & \shortstack[r]{$-8.0$\\$[-16.0, 0.0]$} & \shortstack[r]{$-8.0$\\$[-13.9, -2.5]$} & \shortstack[r]{$-9.9$\\$[-13.2, -6.5]$}\\[2pt]
Prefix + current & \shortstack[r]{$-9.3$\\$[-16.6, -2.0]$} & \shortstack[r]{$-15.7$\\$[-22.9, -9.0]$} & \shortstack[r]{$-4.3$\\$[-12.3, 3.7]$} & \shortstack[r]{$-9.5$\\$[-15.4, -3.5]$} & \shortstack[r]{$-10.0$\\$[-13.5, -6.5]$}\\[2pt]
Last 2 & \shortstack[r]{$-7.3$\\$[-13.7, -1.0]$} & \shortstack[r]{$-3.8$\\$[-8.6, 1.0]$} & \shortstack[r]{$1.2$\\$[-6.7, 9.2]$} & \shortstack[r]{$-3.0$\\$[-9.0, 2.5]$} & \shortstack[r]{$-3.5$\\$[-6.7, -0.3]$}\\[2pt]
Last 4 & \shortstack[r]{$-7.3$\\$[-14.6, 0.0]$} & \shortstack[r]{$-4.8$\\$[-12.9, 2.9]$} & \shortstack[r]{$0.6$\\$[-7.4, 8.6]$} & \shortstack[r]{$-4.0$\\$[-9.5, 1.5]$} & \shortstack[r]{$-4.1$\\$[-7.8, -0.5]$}\\[2pt]
Last 8 & \shortstack[r]{$-12.7$\\$[-19.5, -5.9]$} & \shortstack[r]{$-2.9$\\$[-10.0, 4.8]$} & \shortstack[r]{$-1.2$\\$[-9.8, 7.4]$} & \shortstack[r]{$-7.5$\\$[-12.9, -2.0]$} & \shortstack[r]{$-6.3$\\$[-9.9, -2.8]$}\\[2pt]
Last 16 & \shortstack[r]{$-11.7$\\$[-19.0, -4.4]$} & \shortstack[r]{$-10.0$\\$[-17.1, -2.9]$} & \shortstack[r]{$-4.9$\\$[-12.9, 3.1]$} & \shortstack[r]{$-9.5$\\$[-15.4, -3.5]$} & \shortstack[r]{$-9.2$\\$[-12.7, -5.6]$}\\[2pt]
Full-history control & \shortstack[r]{$-15.1$\\$[-21.5, -8.8]$} & \shortstack[r]{$-16.2$\\$[-23.3, -9.0]$} & \shortstack[r]{$-8.0$\\$[-16.0, 0.0]$} & \shortstack[r]{$-12.4$\\$[-18.4, -6.5]$} & \shortstack[r]{$-13.2$\\$[-16.6, -9.6]$}\\[2pt]
Shuffled & \shortstack[r]{$-14.1$\\$[-21.5, -7.3]$} & \shortstack[r]{$-17.6$\\$[-25.2, -10.5]$} & \shortstack[r]{$-4.3$\\$[-12.3, 3.7]$} & \shortstack[r]{$-13.9$\\$[-19.9, -8.5]$} & \shortstack[r]{$-13.0$\\$[-16.3, -9.4]$}\\[2pt]
Swapped, last 4 & \shortstack[r]{$-8.3$\\$[-15.1, -1.5]$} & \shortstack[r]{$2.4$\\$[-4.3, 9.5]$} & \shortstack[r]{$-1.8$\\$[-9.2, 5.5]$} & \shortstack[r]{$2.0$\\$[-4.0, 8.5]$} & \shortstack[r]{$-1.4$\\$[-4.9, 2.2]$}\\[2pt]
Swapped, full & \shortstack[r]{$-16.6$\\$[-23.4, -9.8]$} & \shortstack[r]{$-15.2$\\$[-22.4, -8.6]$} & \shortstack[r]{$-7.4$\\$[-16.0, 0.6]$} & \shortstack[r]{$-9.5$\\$[-15.9, -3.5]$} & \shortstack[r]{$-12.5$\\$[-16.0, -8.9]$}\\[2pt]
Filler current & \shortstack[r]{$-9.8$\\$[-16.1, -3.4]$} & \shortstack[r]{$-8.6$\\$[-15.7, -1.4]$} & \shortstack[r]{$-3.7$\\$[-11.7, 4.3]$} & \shortstack[r]{$-11.9$\\$[-17.9, -6.0]$} & \shortstack[r]{$-8.7$\\$[-12.2, -5.3]$}\\[2pt]
Token-matched noise & \shortstack[r]{$-13.7$\\$[-20.5, -7.3]$} & \shortstack[r]{$-6.7$\\$[-14.3, 1.0]$} & \shortstack[r]{$-2.5$\\$[-10.4, 5.5]$} & \shortstack[r]{$-7.0$\\$[-13.9, 0.0]$} & \shortstack[r]{$-7.7$\\$[-11.2, -3.9]$}\\[2pt]
Expert + tail & \shortstack[r]{$-16.6$\\$[-22.9, -10.2]$} & \shortstack[r]{$-13.3$\\$[-21.0, -5.7]$} & \shortstack[r]{$-4.9$\\$[-12.9, 3.1]$} & \shortstack[r]{$-13.9$\\$[-19.9, -8.5]$} & \shortstack[r]{$-12.6$\\$[-16.0, -9.0]$}\\[2pt]
Route only & \shortstack[r]{$-9.7$\\$[-20.4, 0.0]$} & \shortstack[r]{$-8.0$\\$[-16.8, 0.8]$} & \shortstack[r]{$1.1$\\$[-10.1, 12.4]$} & \shortstack[r]{$-7.5$\\$[-17.2, 2.2]$} & \shortstack[r]{$-6.4$\\$[-11.4, -1.2]$}\\[2pt]
Own, length-matched & \shortstack[r]{$-14.1$\\$[-21.0, -7.3]$} & \shortstack[r]{$-15.7$\\$[-23.3, -9.0]$} & \shortstack[r]{$-6.7$\\$[-14.7, 1.2]$} & \shortstack[r]{$-11.9$\\$[-17.9, -6.5]$} & \shortstack[r]{$-12.5$\\$[-15.9, -8.7]$}\\[2pt]

\bottomrule
\end{tabular}
\end{table}
\FloatBarrier

\subsection{History Substitution}
We replace only the content of the history, keeping its length fixed.
At the $779$ delayed states, shuffling its (frame, action) pairs leaves a deficit of $13.0$ points ($95\%$ CI $[9.4, 16.3]$), and swapping it for a full-length history from another scene leaves $12.5$ points ($[8.9, 16.0]$), against $13.2$ points for the full-length retained history.
Relative to the full-length retained history, both stay within $5$ points when pooled and $8$ points per policy; these equivalence margins were set after these estimates had been read.
On route, the same substitutions lower success for every policy by $5$--$11$ and $6$--$20$ points relative to the full-length retained history, whereas after a deviation they leave the deficit nearly unchanged.
Filling the same length with copies of the current frame or with noise still leaves deficits of $8.7$ and $7.7$ points.

\subsection{Action Continuation}
Continuation is scored at states where the previously executed action is a forward step.
A decision continues when its greedy action chunk contains no turn, so a chunk that ends in a stop without turning also continues.
Scoring such chunks as non-continuing lowers the retained-history rate from $48.1\%$ to $44.9\%$ at onset and from $46.8\%$ to $41.8\%$ at delayed states, and leaves clearing and the expert unchanged.
The shortest-path expert is queried at the same pose and continues when its first actions, as many as the retained-history chunk holds, contain no turn.
States where the expert returns no action, all within the $3$\,m success radius, are excluded.
Shuffled and swapped histories differ from the full-length retained history by less than $4$ points in pooled point estimates ($50.4\%$ and $51.3\%$ on route, $51.7\%$ and $45.5\%$ at the onset, $48.9\%$ and $47.2\%$ at the delayed states).
Per-policy differences vary in sign, but for every policy and population both stay above clearing and the expert.
Continuation intervals resample episodes separately within each population.
Table~\ref{tab:continuation-per-policy} reports the rates per policy.

\begin{table}[!htbp]
\centering
\caption{Continuation rates per policy, in percent, with $95\%$ episode-bootstrap intervals. States follow a forward step, and states where the expert returns no action are excluded. Pooled rows are the values in Figure~\ref{fig:continuation}. Two on-route states each of NaVILA and Uni-NaVid have no swapped history, leaving $235$ and $316$ states ($942$ pooled).}
\label{tab:continuation-per-policy}
\small
\setlength{\tabcolsep}{3pt}
\begin{tabular}{@{}lrrrrrr@{}}
\toprule
Policy & $n$ & Retained & Cleared & Shuffled & Swapped & Expert\\
\midrule
\multicolumn{7}{@{}l}{\textit{On route}}\\
NaVIDA & 172 & 39.0 [32.0, 45.9] & 22.7 [16.3, 29.1] & 39.5 [32.6, 47.1] & 37.8 [30.8, 45.3] & 32.0 [25.0, 38.4]\\
StreamVLN & 219 & 36.5 [30.1, 42.9] & 20.1 [15.1, 25.6] & 39.7 [33.3, 46.1] & 37.0 [30.6, 43.4] & 32.4 [26.5, 38.8]\\
NaVILA & 237 & 76.8 [71.3, 82.3] & 18.1 [13.5, 23.2] & 51.5 [45.6, 57.8] & 68.1 [62.1, 74.0] & 35.4 [29.1, 41.8]\\
Uni-NaVid & 318 & 43.7 [38.1, 49.1] & 21.1 [17.0, 25.8] & 62.9 [57.2, 68.2] & 56.0 [50.3, 61.4] & 26.1 [21.4, 30.8]\\
Pooled & 946 & 49.5 [46.2, 52.7] & 20.4 [17.9, 22.8] & 50.4 [47.3, 53.6] & 51.3 [48.3, 54.6] & 31.0 [28.1, 33.9]\\
\midrule
\multicolumn{7}{@{}l}{\textit{Onset}}\\
NaVIDA & 160 & 35.0 [27.5, 42.5] & 11.9 [7.5, 16.9] & 40.6 [32.5, 48.8] & 24.4 [18.1, 31.2] & 1.2 [0.0, 3.1]\\
StreamVLN & 169 & 33.1 [26.0, 40.2] & 13.6 [8.9, 18.9] & 38.5 [31.9, 46.2] & 33.1 [26.0, 40.2] & 1.8 [0.0, 4.1]\\
NaVILA & 182 & 68.7 [62.1, 75.3] & 5.5 [2.2, 9.3] & 48.4 [40.7, 55.5] & 56.0 [48.4, 63.7] & 1.1 [0.0, 2.7]\\
Uni-NaVid & 215 & 52.1 [45.6, 59.1] & 12.6 [8.4, 17.2] & 73.0 [67.0, 79.1] & 61.9 [55.3, 68.4] & 3.3 [1.4, 5.6]\\
Pooled & 726 & 48.1 [44.5, 51.8] & 10.9 [8.8, 13.4] & 51.7 [47.9, 55.2] & 45.5 [41.9, 49.2] & 1.9 [1.0, 3.0]\\
\midrule
\multicolumn{7}{@{}l}{\textit{Delayed}}\\
NaVIDA & 119 & 40.3 [31.9, 49.6] & 16.0 [10.1, 22.7] & 39.5 [31.1, 48.7] & 28.6 [21.0, 37.0] & 3.4 [0.8, 6.7]\\
StreamVLN & 133 & 33.8 [26.3, 42.1] & 12.8 [7.5, 18.8] & 36.8 [29.3, 45.1] & 36.8 [28.6, 45.1] & 2.3 [0.0, 5.3]\\
NaVILA & 110 & 60.0 [50.9, 69.1] & 1.8 [0.0, 4.5] & 48.2 [39.1, 57.3] & 63.6 [54.5, 72.7] & 5.5 [1.8, 10.0]\\
Uni-NaVid & 176 & 52.8 [46.0, 60.2] & 6.8 [3.4, 10.8] & 64.8 [58.0, 72.2] & 57.4 [50.0, 64.8] & 7.4 [4.0, 11.4]\\
Pooled & 538 & 46.8 [42.8, 51.3] & 9.3 [6.9, 11.9] & 48.9 [44.8, 53.0] & 47.2 [43.1, 51.5] & 4.8 [3.2, 6.7]\\
\bottomrule
\end{tabular}
\end{table}
\FloatBarrier

\subsection{Greedy Selection and Sampling Coverage}
The greedy selection oracle takes the successful continuation whenever either retained or cleared history succeeds at the same onset state. %
On all $915$ onset states it reaches $35.0\%$ success, $6.3$ points above always clearing (Figure~\ref{fig:greedy_selection}): clearing rescues $195$ of the $790$ states where the retained rollout fails and breaks $58$ of the $125$ where it succeeds, and both rollouts fail on $65.0\%$ of the states.
The per-policy bars in Figure~\ref{fig:greedy_selection} show pointwise Wilson intervals computed from the corresponding success counts.
The oracle's $35.0\%$ rate bounds selection between those two frozen greedy continuations, rather than recovery with a trained policy or a larger sampling budget.
Sampling states are drawn with a fixed seed and route-distance stratification from retained-greedy failures, selecting $80$ states per policy.
Each condition produces eight continuations with temperature $1.0$, top-$p=0.95$, and paired random seeds.
For a state with $c$ successful samples among eight, the coverage estimate at budget $k$ is $1-\binom{8-c}{k}/\binom{8}{k}$, taking the numerator as zero when $8-c<k$.
Reported $\mathrm{pass}@k$ averages these estimates over states, with pointwise state-bootstrap intervals.
These offline coverage estimates assume success can be recognized among sampled continuations and are not deployment success rates.
Table~\ref{tab:recovery-policy-coverage} reports policy-level sampling coverage.

\begin{table}[!htbp]
\centering
\caption{Offline coverage after eight samples on retained-greedy failures ($80$ onset states per policy). Values are percentages with $95\%$ state-bootstrap intervals. The final column is the fraction with no successful cleared continuation among eight samples.}
\label{tab:recovery-policy-coverage}
\small
\setlength{\tabcolsep}{5pt}
\begin{tabular}{lrrr}
\toprule
Policy & Retained $\mathrm{pass}@8$ & Cleared $\mathrm{pass}@8$ & Cleared: all fail\\
\midrule
NaVIDA & 36.2 [26.2, 47.5] & 66.2 [56.2, 76.2] & 33.8\\
StreamVLN & 32.5 [22.5, 43.8] & 68.8 [58.8, 78.8] & 31.2\\
NaVILA & 36.2 [26.2, 47.5] & 71.2 [61.3, 80.0] & 28.7\\
Uni-NaVid & 16.2 [8.8, 25.0] & 40.0 [30.0, 51.2] & 60.0\\

\bottomrule
\end{tabular}
\end{table}

\FloatBarrier
\section{Method Details}\label{app:method}

\subsection{Base Policy}
The frozen policy NaVIDA-Qwen3VL follows the official training recipe of NaVIDA~\citep{zhu2026navida} with Qwen3-VL-4B-Instruct as the backbone, and Table~\ref{tab:base-config} lists its configuration.
\begin{table}[h]
\centering
\caption{Training configuration of the base policy NaVIDA-Qwen3VL.}
\label{tab:base-config}
\small
\begin{tabular}{l>{\raggedright\arraybackslash}p{0.66\linewidth}}
\toprule
Configuration & Setting \\
\midrule
Backbone & Qwen3-VL-4B-Instruct \\
Training data & $0.82$M (R2R-CE and RxR-CE), $0.76$M (DAgger) and $3.35$M (ScaleVLN); $4.92$M samples in total (components rounded) \\
Stages & Stage I on R2R-CE and RxR-CE, DAgger collection, Stage II on all data \\
Epochs & $1$ per stage \\
Learning rate & $2\times10^{-5}$, cosine decay \\
Global batch size & $64$ (Stage I), $128$ (Stage II) \\
History frames & $8$, sampled uniformly \\
Image resolution & $352\times288$ \\
Precision and parallelism & BF16, FlashAttention-2, DeepSpeed ZeRO-1 \\
Hardware & $16$ / $24$ / $32$ A100-80GB for Stage I / DAgger / Stage II (about $6$ / $5$ / $18$ hours) \\
\bottomrule
\end{tabular}
\end{table}

\subsection{Judge}
The lookahead set of \eqref{eq:label} is $V_t=\{t+3,t+6,t+9,t+12,t+15\}\cap[0,T)$ in atomic steps, where $T$ is the length of the episode.
It is set to $\emptyset$ when it has fewer than three elements, so near the end of an episode a decision point is labeled LOST exactly when $h_t>S_d$.
The input $E_t$ contains up to nine frames spaced evenly over the history since the episode start, the last being $o_t$, together with the instruction, up to the last twelve executed atomic actions and the chunk $c_t$ written as action names.
The judge answers with one token.
Its prompt also lists a third answer that no training row uses.
The probability of that answer rescales $p_\phi(\mathrm{OK}\mid E_t)$ and $p_\phi(\mathrm{LOST}\mid E_t)$ by the same factor, so $m_t$ is unchanged.
Training rows come from rollouts of $\pi_\theta$ on the training splits of R2R-CE and RxR-CE and on ScaleVLN~\citep{wang2023scalevln}.
On these rollouts, $64.5\%$, $55.4\%$ and $56.0\%$ of the steps that exceed $S_d$ but are labeled OK lie in successful episodes for R2R-CE, RxR-CE and ScaleVLN, against less than $18\%$ of the steps labeled LOST.
Rows are drawn from every failed episode and an equal number of successful ones, at most $24$ per episode, and the two larger datasets are then subsampled to the row count of R2R-CE.
OK and LOST rows are drawn in equal proportion during training.
The adapter $\phi$ is a LoRA of rank $32$ with scaling factor $2$ (LoRA alpha $64$) on the query, key, value and output projections of the language model.
Five of the checkpoints saved every $300$ steps ($900$, $1{,}500$, $2{,}100$, $2{,}700$ and $3{,}600$) are compared by episode-level recall at $\alpha=5\%$ on the selection half described below, and step $1{,}500$ is used.
The judge scores every decision point whose chunk contains no stop, with the window $W=3$.

\subsection{Calibration}
The judge's training excludes $15\%$ of the reference paths of the training splits, chosen by a hash of the path, and every scene contributes to both sides.
The held-out paths are split into a selection half and a calibration half.
For each dataset, the calibration episodes are the rollouts of $\pi_\theta$ on the calibration half that it completes with greedy decoding, and each dataset receives its own $\tau$.
The score $Z_e$ is the maximum of $\bar m_t$ over all decision points of the episode whose chunk does not open with a stop, with the same window as at deployment, and $\tau=+\infty$ when $\lceil(n+1)(1-\alpha)\rceil>n$.
Unlike deployment, calibration also scores the final decision point when its chunk contains a stop without opening with one; since that chunk ends the episode, this can only raise $Z_e$ and hence $\tau$, and so does not loosen the bound of Section~\ref{sec:method:judge}.
The simulator forces a stop once the distance to the goal has stayed unchanged for more than $25$ consecutive atomic actions or more than $400$ atomic actions have been executed.
At a forced stop, the chunk scored at deployment can differ from the logged chunk used for calibration, so the bound of Section~\ref{sec:method:judge} can be exceeded by at most the probability that a completed episode ends in a forced stop.
Forced stops end $0.15\%$ and $0.27\%$ of the successful rollouts of $\pi_\theta$ on the training splits of R2R-CE and RxR-CE.
We use $\alpha=10\%$ for R2R-CE ($n=635$, $\tau=2.54$) and $\alpha=7.5\%$ for RxR-CE ($n=1{,}103$, $\tau=3.33$).
These levels gave the highest closed-loop success among the levels we evaluated with the recovery policy on the val-unseen splits, four for R2R-CE and three for RxR-CE.
They were therefore selected on the evaluation splits.
Figure~\ref{fig:method} shows R2R-CE val-unseen episode $959$, rerun with the threshold calibrated at $\alpha=7.5\%$ ($\tau=3.08$) instead of the $\alpha=10\%$ used for R2R-CE.

\subsection{Training Start States}
The judge is run offline on every decision point whose chunk does not open with a stop, over rollouts of $\pi_\theta$ on the training splits of R2R-CE and RxR-CE and on ScaleVLN, with $W=3$ and per-dataset thresholds calibrated at $\alpha=5\%$ ($\tau=3.83$, $3.71$ and $3.08$ for R2R-CE, RxR-CE and ScaleVLN).
The first alarm of each episode becomes a start state, and episodes on the judge's held-out paths are excluded.
The pool holds $5{,}510$ states.
Of these, $24.2\%$ come from episodes that $\pi_\theta$ completes, and $25.8\%$ lie within $S_d$ of the route.
Because the offline run also scores chunks that contain but do not open with a stop, $283$ of these states lie at a final decision point at which the deployed judge is not queried.
A training rollout places the agent at the recorded pose with the remaining step budget, seeds the history with the current frame alone, and runs without the judge.

\subsection{Counterfactual Stop Advantage}
Training also uses a stop term. An outcome reward cannot tell at which chunk a rollout should have stopped, and because stops are rare, a group seldom contains both choices at one state.
A stop ends the episode, however, so its return has a closed form.
Let $c_{i,j}$ be the $j$-th chunk of $\zeta_i$, $R^{\mathrm{stop}}_{i,j}$ the reward of \eqref{eq:reward} had $\zeta_i$ stopped at the start of $c_{i,j}$, and $p_{i,j}$ the probability of opening $c_{i,j}$ with a stop under the policy that sampled $\zeta_i$.
We split the regret of not stopping between the two choices:
\begin{equation}
\Delta_{i,j}=R^{\mathrm{stop}}_{i,j}-R_i,
\qquad
A^{\mathrm{stop}}_{i,j}=(1-p_{i,j})\,\Delta_{i,j},
\qquad
A^{\mathrm{cont}}_{i,j}=-p_{i,j}\,\Delta_{i,j}.
\label{eq:stop}
\end{equation}
The pair has zero mean under that policy, $p_{i,j}A^{\mathrm{stop}}_{i,j}+(1-p_{i,j})A^{\mathrm{cont}}_{i,j}=0$, so its two terms need no relative weight, and they share one per-chunk weight.
The advantage of $c_{i,j}$ adds $A^{\mathrm{cont}}_{i,j}$ to $\hat A_i$, and the stop token receives $A^{\mathrm{stop}}_{i,j}$.
Removing this term has no measurable effect on success or on stopping behavior in our ablation (SR $66.3$ against $66.2$ with the term at iteration 3 and $66.6$ against $67.0$ at iteration 9, $p=0.92$ and $0.53$), so the main text does not treat it as a component.

\subsection{Recovery Objective}
The objective minimized over $\psi$ is
\begin{equation}
\mathcal{L}(\psi)=\frac{1}{N}\sum_{i,j} w_{i,j}\Bigl[\ell\bigl(c_{i,j},\,\hat A_i+\omega_{i,j}A^{\mathrm{cont}}_{i,j}\bigr)+\frac{1}{K_{i,j}}\,\ell\bigl(\mathrm{stop},\,\omega_{i,j}A^{\mathrm{stop}}_{i,j}\bigr)+\beta\,\hat D_{i,j}\Bigr],
\label{eq:objective}
\end{equation}
where $\ell(x,A)$ is the clipped PPO loss for advantage $A$, averaged over the tokens of $x$; without the two stop terms it reduces to \eqref{eq:grpo}.
For a chunk these are the unmasked tokens of its full generated text, $K_{i,j}$ counts them for $c_{i,j}$, and the mask removes sub-actions that parse to no atomic action and therefore execute a uniformly random movement.
The ratio of $\pi_{\theta,\psi}$ to the sampling policy is clipped to $[0.8,1.28]$, with a dual clip at $3$ when $A<0$.
The term $\hat D_{i,j}$ estimates $\KL(\pi_{\theta,\psi}\,\|\,\pi_\theta)$ on the same unmasked tokens by $\frac{\pi_\theta}{\pi_{\theta,\psi}}-\log\frac{\pi_\theta}{\pi_{\theta,\psi}}-1$, and $\beta=0.04$.
There is no KL-based early stopping.
The weight $w_{i,j}\propto 1/J_i$, where $J_i$ counts the chunks of $\zeta_i$ used for training, is normalized to mean one over the chunks each worker collects in an iteration, so rollouts on one worker receive the same total weight.
The weight $\omega_{i,j}$ is one for chunks that start within $5$\,m of the goal.
Beyond $5$\,m, up to six chunks per rollout are drawn at random, two from each of the ranges $[5,10)$, $[10,15)$ and $[15,\infty)$\,m where available and the rest from the remaining far chunks; they share one unit of weight, and the others get zero.
This keeps the many far chunks, where $R^{\mathrm{stop}}_{i,j}=0$, from repeating the same signal.
Because $\omega_{i,j}$ scales both terms of \eqref{eq:stop} alike, the pair keeps its zero mean.
Here $N$ is the number of chunks across all workers in one optimizer step, and the stop term is dropped for a chunk that opens with a stop.
The stop probability $p_{i,j}$ is read from the sampling policy at the first generated position and the sampling temperature.
All likelihoods in \eqref{eq:objective} and the stop probability $p_{i,j}$ are computed from temperature-scaled checkpoint logits without the repetition penalty of $1.05$ used to sample rollouts.
The standard deviation in \eqref{eq:advantage} uses Bessel's correction.

\subsection{Training and Decoding}
Each iteration draws $128$ alarm states and samples $G=8$ rollouts per state at temperature $1.6$ with top-$p=1$.
The reward uses $\lambda=0.3$ and the environment's $3$\,m success radius, and $L_i$ sums the displacements between the poses recorded after each step, which leaves out the displacement of the first step of the suffix; $R^{\mathrm{stop}}_{i,j}$ uses the same convention.
Optimization uses $16$ samples per step and a learning rate that decays linearly from $5\times10^{-5}$ toward $3\times10^{-5}$.
The adapter $\psi$ is a zero-initialized LoRA of rank $16$ with scaling factor $2$ (LoRA alpha $32$) on the query, key, value and output projections of the language model, on the attention of four vision blocks and on the vision merger, so $\pi_{\theta,\psi}=\pi_\theta$ when training starts.
Decoding is greedy before and after the alarm, at deployment and in every closed-loop evaluation.

\FloatBarrier
\section{Additional Results}\label{app:results}

\subsection{Training-Side Ablation over Iterations}\label{app:train-iters}
Table~\ref{tab:ablation-train-full} extends Table~\ref{tab:ablation-training} to all evaluated iterations.
On the $527$ alarmed R2R-CE episodes, the greedy success of $47.1\%$ at iteration 15 (Section~\ref{sec:exp:ablation}) includes $60$ of the $205$ base failures on which all eight samples had failed; eight samples give only a lower bound on what sampling reaches.
Without the judge, GRPO$_{\mathrm{start}}$ stays at the base level ($62.3$, $p=0.21$).
A binary success reward in place of suffix-SPL gives $66.5$ at iteration 9, against $67.0$ ($p=0.39$).

\begin{table}[h]
\centering
\caption{Training-side ablation on R2R-CE: SR at iterations 3, 9, 15 and 19, counted from 0. Start: where rollouts or demonstrations begin. Clear: whether they begin from cleared history; the row without it also retains the history at deployment. $^*$No judge at deployment. GRPO$_{\mathrm{binary}}$: binary success reward instead of suffix-SPL. --: does not apply or was not run.}
\label{tab:ablation-train-full}
\small
\setlength{\tabcolsep}{4pt}\setlength{\aboverulesep}{0pt}\setlength{\belowrulesep}{0pt}\setlength{\extrarowheight}{0.1ex}
\begin{tabular}{lcccccc}
\toprule
\multirow{2}{*}{Training} & \multirow{2}{*}{Start} & \multirow{2}{*}{Clear} & \multicolumn{4}{c}{SR$\uparrow$ at iteration}\\
\cmidrule(lr){4-7}
 & & & 3 & 9 & 15 & 19 \\
\midrule
None (clear only) & -- & -- & \multicolumn{4}{c}{65.6} \\
SFT & alarm & \cm & 61.8 & 62.9 & -- & -- \\
GRPO$_{\mathrm{start}}^*$ & step 0 & -- & -- & 62.3 & -- & -- \\
GRPO$_{\mathrm{start}}$ & step 0 & -- & -- & 66.2 & -- & -- \\
GRPO$_{\mathrm{keep}}$ & alarm & \xm & 65.5 & 66.2 & 67.4 & 67.8 \\
GRPO$_{\mathrm{binary}}$ & alarm & \cm & 67.2 & 66.5 & -- & -- \\
\rowcolor{blue!10}
GRPO (ours) & alarm & \cm & 66.2 & 67.0 & 69.0 & 68.1 \\
\bottomrule
\end{tabular}
\end{table}

\subsection{Judge Counts}\label{app:judge-counts}
Table~\ref{tab:judge-counts} gives the counts behind Figure~\ref{fig:judge_outcomes}.
The false-alarm rates of $13.0\%$ and $8.0\%$ compare with nominal levels of $10\%$ and $7.5\%$; the calibration episodes come from held-out paths in the training scenes rather than from unseen scenes, so the rate can exceed the nominal level, as on R2R-CE.
The other $43.7\%$ of R2R-CE failures never receive an alarm, so detection bounds the gain.
As in Section~\ref{sec:analysis:recovery}, clearing at an alarm can break an episode as well as rescue one: among alarmed episodes, the recovery policy raises the share of failures it completes from $29.1\%$ with clearing alone to $43.7\%$ and lowers the share of successes it breaks from $50.0\%$ to $44.7\%$ on R2R-CE.
Rescued and broken episodes would cancel only if base successes made up $49.4\%$ of the alarmed episodes on R2R-CE and $51.3\%$ on RxR-CE, against the observed $28.8\%$ and $24.4\%$, and allowing a second clearing per episode adds nothing.
Alarms come a median of $13$ atomic steps after the deviation onset $t_d$ on R2R-CE and $21$ on RxR-CE, partly because the window average of \eqref{eq:score} delays the score, and some alarms come before $t_d$.
The random trigger of Table~\ref{tab:ablation-trigger} alarms on base failures only at the base failure rate ($36.6\%$ against $36.2\%$).

\begin{table}[h]
\centering
\begin{minipage}[t]{0.62\textwidth}
\centering
\caption{Counts behind Figure~\ref{fig:judge_outcomes}. Brackets: Wilson $95\%$ intervals. RxR-CE recall includes 99 base failures whose goal lies on a disconnected part of the navigation mesh, and is 589/1{,}187 without them.}
\label{tab:judge-counts}
\small
\setlength{\tabcolsep}{2.6pt}\setlength{\aboverulesep}{0pt}\setlength{\belowrulesep}{0pt}\setlength{\extrarowheight}{0.1ex}
\begin{tabular}{lcc}
\toprule
 & R2R-CE ($\alpha=10\%$) & RxR-CE ($\alpha=7.5\%$) \\
\midrule
Alarm & 527/1{,}839 & 779/3{,}669 \\
Prec. & 375/527 & 589/779 \\
Recall & 375/666 & 589/1{,}286 \\
False alarm & 152/1{,}173 & 190/2{,}383 \\
Rescued & 164/375 [38.8, 48.8] & 317/589 [49.8, 57.8] \\
Broken & 68/152 [37.1, 52.7] & 97/190 [44.0, 58.1] \\
\bottomrule
\end{tabular}
\end{minipage}
\end{table}

\subsection{Alarm Level}\label{app:alarm-level}
Table~\ref{tab:alpha-sweep} varies the level $\alpha$ at which the threshold $\tau$ is calibrated, with the recovery policy of the main results.
Every level improves on the base, and the reported levels give the highest success on both datasets.
In leave-one-scene-out cross-validation, which selects $\alpha$ on ten val-unseen scenes and scores the eleventh, ten of the eleven folds choose the reported level.
The false-alarm rate follows $\alpha$ and exceeds it by at most $3.0$ points on R2R-CE and $0.6$ on RxR-CE.

\begin{table}[h]
\centering
\caption{Alarm level. Alarms: share of all episodes; false alarms: share of base successes. $^\dagger$Reported level.}
\label{tab:alpha-sweep}
\small
\setlength{\tabcolsep}{4pt}
\begin{tabular}{lcccccc}
\toprule
Dataset & $\alpha$ & $\tau$ & Alarms & False alarms & SR$\uparrow$ & SPL$\uparrow$ \\
\midrule
\multirow{5}{*}{R2R-CE} & 5.0\% & 3.83 & 16.3 & 5.5 & 66.9 & 60.3 \\
 & 7.5\% & 3.08 & 23.3 & 9.4 & 67.9 & 60.4 \\
 & 10.0\%$^\dagger$ & 2.54 & 28.7 & 13.0 & 69.0 & 62.3 \\
 & 12.5\% & 2.29 & 30.7 & 14.7 & 67.8 & 59.7 \\
 & Base & -- & -- & -- & 63.8 & 59.0 \\
\midrule
\multirow{4}{*}{RxR-CE} & 2.5\% & 4.37 & 11.2 & 3.1 & 67.0 & 56.7 \\
 & 7.5\%$^\dagger$ & 3.33 & 21.2 & 8.0 & 70.9 & 60.8 \\
 & 12.5\% & 2.71 & 26.7 & 11.7 & 67.8 & 56.4 \\
 & Base & -- & -- & -- & 64.9 & 55.9 \\
\bottomrule
\end{tabular}
\end{table}

\subsection{Judge Query Interval}
The deployed loop queries the judge at every decision point whose chunk contains no stop.
Querying it at every second or fourth such decision point raises about as many alarms but later, and SR falls to $67.9$ and $67.0$ (Table~\ref{tab:query-interval}).

\begin{table}[h]
\centering
\caption{Judge query interval on R2R-CE. $^\dagger$Deployed interval.}
\label{tab:query-interval}
\small
\setlength{\tabcolsep}{6pt}
\begin{tabular}{lccc}
\toprule
Interval (decision points) & SR$\uparrow$ & SPL$\uparrow$ & Alarms \\
\midrule
1$^\dagger$ & 69.0 & 62.3 & 527 \\
2 & 67.9 & 59.9 & 535 \\
4 & 67.0 & 58.8 & 540 \\
\bottomrule
\end{tabular}
\end{table}

\subsection{Zero-Shot Object Navigation}
We evaluate on the HM3D-OVON val-unseen split under its official protocol~\citep{yokoyama2024hm3dovon}, as for the compared methods in Table~\ref{tab:ovon}, with $30^\circ$ turns and success when the agent stops within $1$\,m of an instance of the goal category within $500$ steps.
The goal category is given to the policy as its instruction, \texttt{Walk to the \textless{}category\textgreater{} and stop by it.}, in the prompt of Appendix~\ref{app:prompts}.
Neither the base nor the two adapters are trained on object-goal data, and both adapters run unchanged, with the threshold calibrated on R2R-CE.
The route enters only the training labels of \eqref{eq:label}, and the judge reads only what is available to the agent (Section~\ref{sec:method:judge}), so it can run on a task without a route.
On HM3D-OVON the labels of \eqref{eq:label} are undefined, so an alarm marks a state whose deviation score exceeds the R2R-CE threshold, not a measured deviation.
The bound of Section~\ref{sec:method:judge} does not carry over either, because these episodes are not exchangeable with the R2R-CE calibration episodes, and the threshold acts as a fixed operating point.
We therefore evaluate the alarm only through the SR and SPL of the full system (Table~\ref{tab:ovon}), against $40.7$ and $21.7$ for the base under the same protocol.

\begin{table}[h]
\centering
\caption{Comparison on the HM3D-OVON val-unseen split. Results of other methods are taken from their papers, except VLFM, whose results are from \citet{yokoyama2024hm3dovon}. \textbf{Bold} and \underline{underline} mark the best and second-best results. $^*$Zero-shot evaluation, without training on HM3D-OVON. VLFM, DAgRL+OD and MTU3D also use depth and odometry, ABot-N0 uses panoramic RGB and NavFoM four cameras. $^\dagger$360 randomly sampled episodes.}
\label{tab:ovon}
\small
\setlength{\tabcolsep}{4pt}\setlength{\aboverulesep}{0pt}\setlength{\belowrulesep}{0pt}\setlength{\extrarowheight}{0.1ex}
\begin{tabular}{lcc}
\toprule
Method & SR$\uparrow$ & SPL$\uparrow$\\
\midrule
DAgRL+OD~\pub{IROS24}~\citep{yokoyama2024hm3dovon} & 37.1 & 19.9 \\
MTU3D$^\dagger$~\pub{ICCV25}~\citep{zhu2025mtu3d} & 40.8 & 12.1 \\
ABot-N0~\pub{arXiv26}~\citep{chu2026abotn0} & \underline{54.0} & 30.5 \\
Qwen-RobotNav-4B~\pub{arXiv26}~\citep{qwenrobotnav} & 53.1 & 20.9 \\
Qwen-RobotNav-8B~\pub{arXiv26}~\citep{qwenrobotnav} & 51.2 & 24.0 \\
\midrule
VLFM$^*$~\pub{ICRA24}~\citep{yokoyama2024vlfm} & 35.2 & 19.6 \\
Uni-NaVid$^*$~\pub{RSS25}~\citep{zhang2024uninavid} & 39.5 & 19.8 \\
NavFoM$^*$~\pub{ICLR26}~\citep{zhang2026navfom} & 45.2 & \underline{31.9} \\
JanusVLN$^*$~\pub{ICLR26}~\citep{zeng2026janusvln} & 44.9 & 31.7 \\
\midrule
\rowcolor{blue!10}
\method{} (Ours)$^*$ & \textbf{57.2} & \textbf{35.4} \\
\bottomrule
\end{tabular}
\end{table}

\FloatBarrier
\clearpage
\section{Qualitative Examples}\label{app:qualitative}
The episode of Figure~\ref{fig:qualitative} and the six supplementary figures, three per dataset with three episodes each, were selected by hand among episodes that the frozen policy fails and \method{} completes, so they illustrate recoveries rather than typical outcomes; Table~\ref{tab:judge-counts} gives how often an alarm rescues or breaks an episode.

\begin{figure}[!htbp]
\centering
\includegraphics[width=\linewidth,height=0.82\textheight,keepaspectratio]{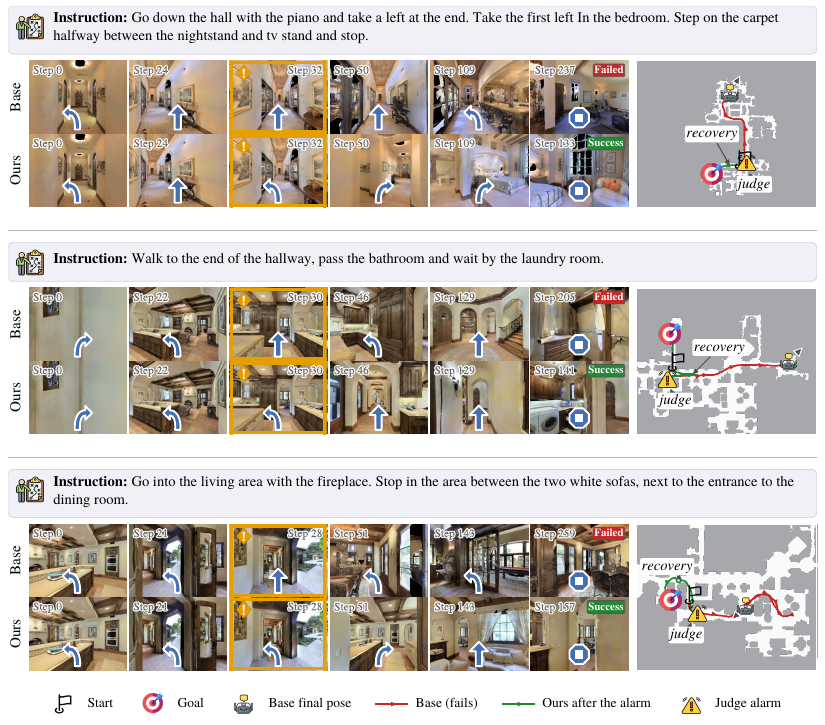}
\caption{Additional R2R-CE episodes (1 of 3) in which the frozen policy fails and \method{} succeeds, shown as in Figure~\ref{fig:qualitative}.}
\label{fig:qualitative_supp_r2r_1}
\end{figure}

\begin{figure}[p]
\centering
\includegraphics[width=\linewidth,height=0.82\textheight,keepaspectratio]{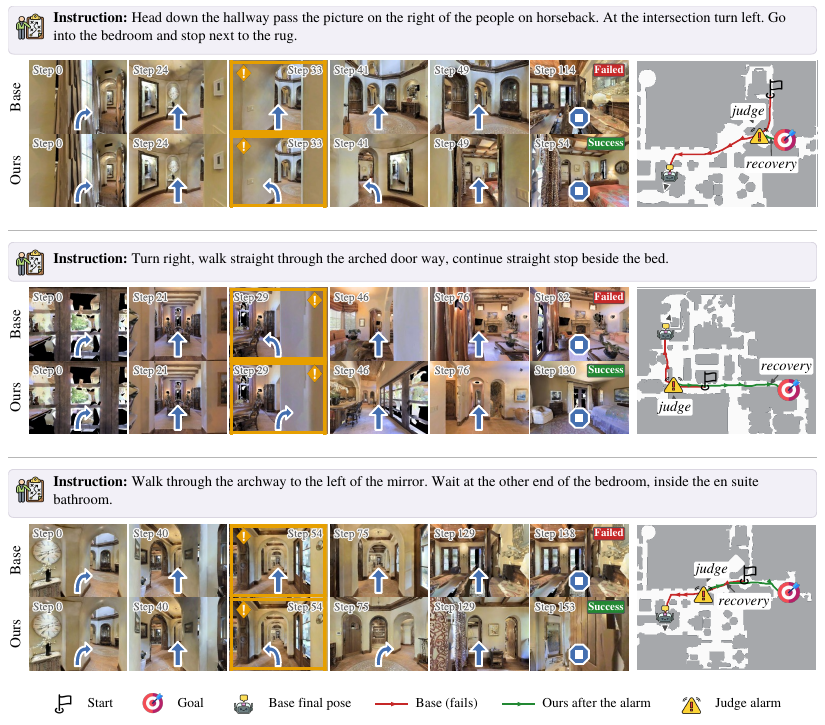}
\caption{Additional R2R-CE episodes (2 of 3), shown as in Figure~\ref{fig:qualitative}.}
\label{fig:qualitative_supp_r2r_2}
\end{figure}

\begin{figure}[p]
\centering
\includegraphics[width=\linewidth,height=0.82\textheight,keepaspectratio]{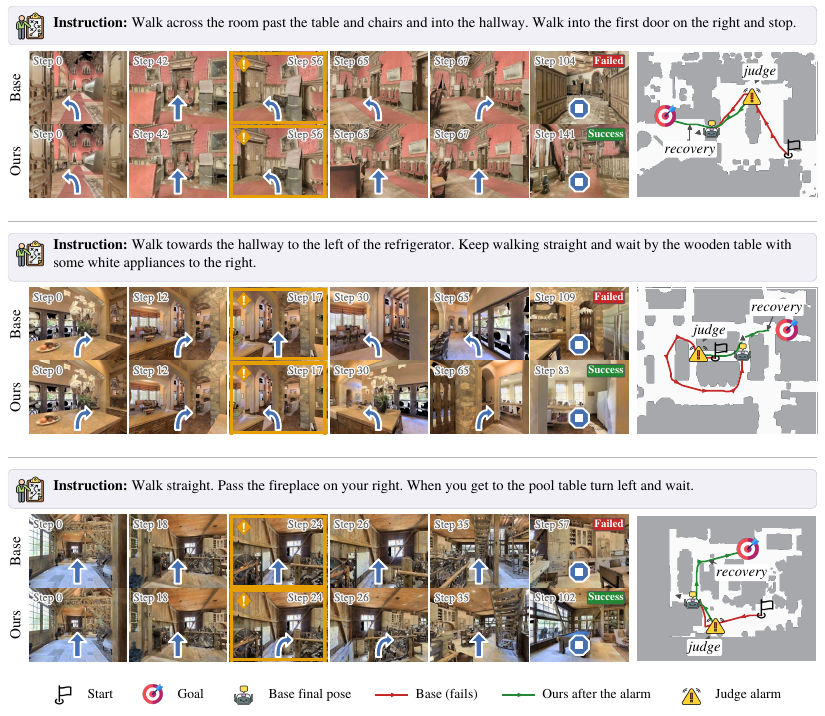}
\caption{Additional R2R-CE episodes (3 of 3), shown as in Figure~\ref{fig:qualitative}.}
\label{fig:qualitative_supp_r2r_3}
\end{figure}

\begin{figure}[p]
\centering
\includegraphics[width=\linewidth,height=0.82\textheight,keepaspectratio]{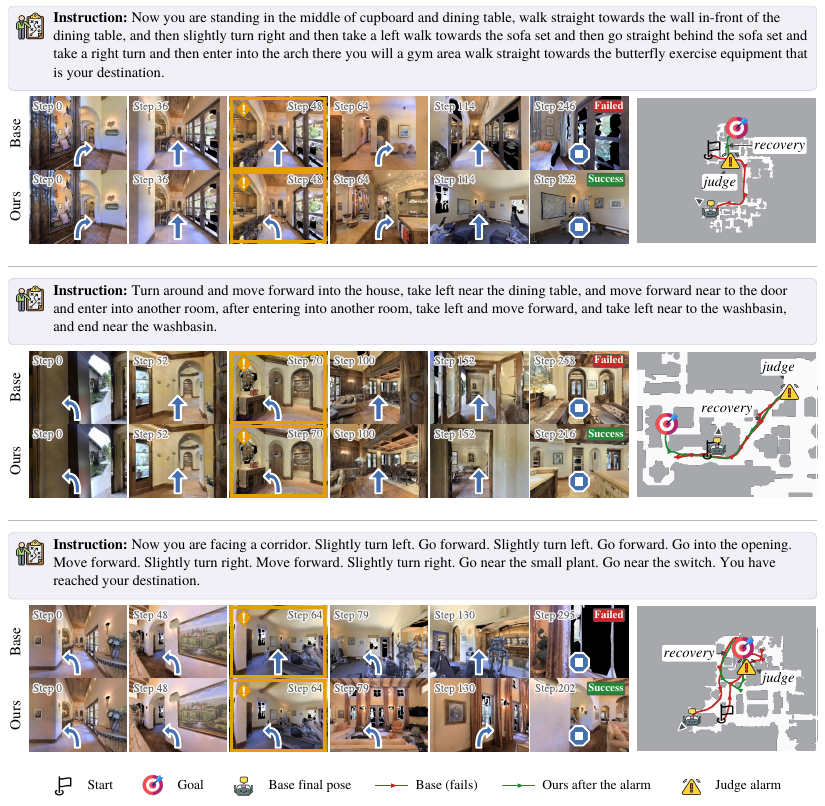}
\caption{Additional RxR-CE episodes (1 of 3) in which the frozen policy fails and \method{} succeeds, shown as in Figure~\ref{fig:qualitative}.}
\label{fig:qualitative_supp_rxr_1}
\end{figure}

\begin{figure}[p]
\centering
\includegraphics[width=\linewidth,height=0.82\textheight,keepaspectratio]{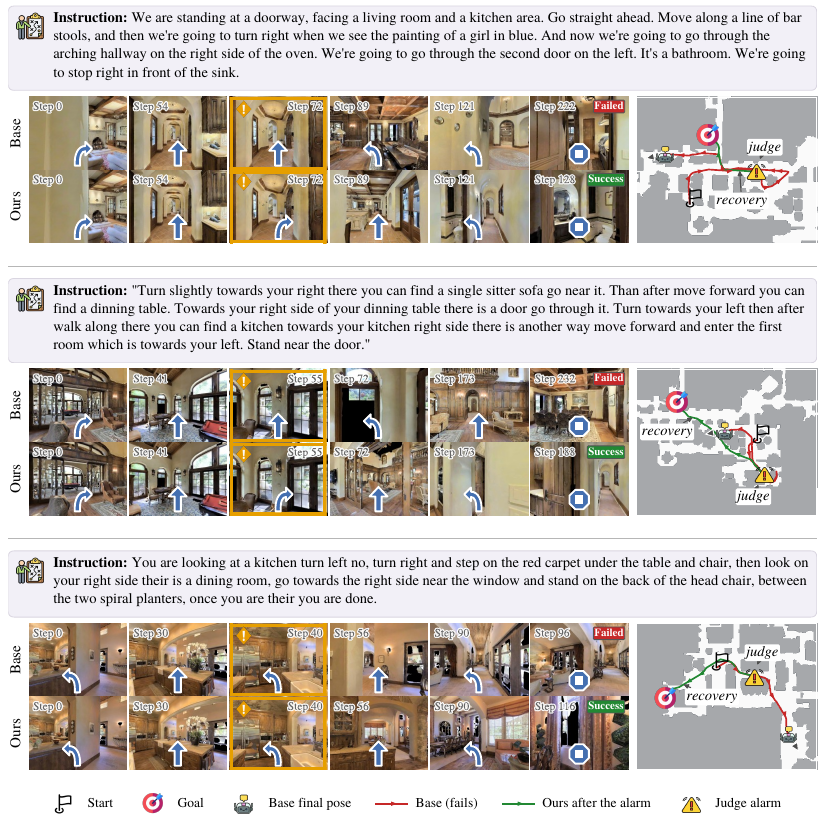}
\caption{Additional RxR-CE episodes (2 of 3), shown as in Figure~\ref{fig:qualitative}.}
\label{fig:qualitative_supp_rxr_2}
\end{figure}

\begin{figure}[p]
\centering
\includegraphics[width=\linewidth,height=0.82\textheight,keepaspectratio]{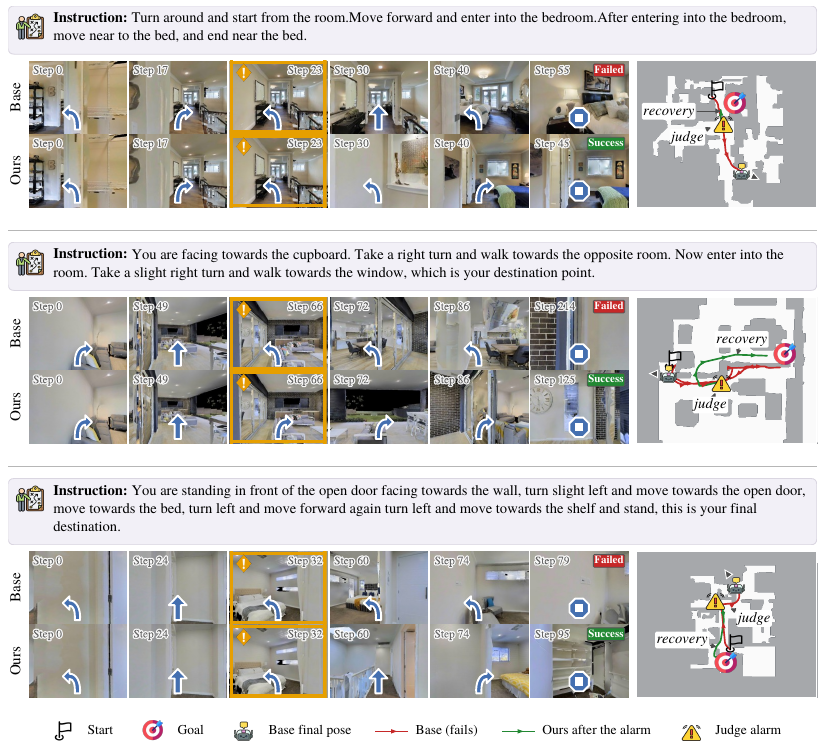}
\caption{Additional RxR-CE episodes (3 of 3), shown as in Figure~\ref{fig:qualitative}.}
\label{fig:qualitative_supp_rxr_3}
\end{figure}

\FloatBarrier
\section{Prompts}\label{app:prompts}

Angle brackets mark inputs filled in at run time, frames enter as image tokens with no text between them, and within a turn each line of a box is one line of the prompt, with wrapped continuations indented.

\subsection{Navigation Policy}
The frozen base, the recovery policy at deployment and in GRPO training, and the SFT baseline all use the NaVIDA prompt below, with up to eight earlier frames sampled uniformly as \texttt{\textless{}frame 1\textgreater{}}\,\dots\,\texttt{\textless{}frame 8\textgreater{}} and the current frame as \texttt{\textless{}frame 9\textgreater{}}.
\begin{center}
\fbox{\begin{minipage}{0.95\linewidth}
\footnotesize\ttfamily\frenchspacing\raggedright\setlength{\parindent}{0pt}\setlength{\parskip}{1pt}
{\rmfamily\bfseries System}\par
You are a helpful assistant.\par
\smallskip
{\rmfamily\bfseries User}\par
\hangindent=1.5em Imagine you are a robot programmed for navigation tasks. You have been given a video of historical observations\textless{}frame 1\textgreater{}...\textless{}frame 8\textgreater{}and an image of the current observation\textless{}frame 9\textgreater{}. Your assigned task is: \textquotesingle\textless{}instruction\textgreater{}\textquotesingle. Analyze this series of images to decide your next move, which could involve turning left or right by a specific degree or moving forward a certain distance.\par
\smallskip
{\rmfamily\bfseries Assistant}\par
\textless{}answer\textgreater{}
\end{minipage}}
\end{center}
At the first step of an episode and after a clearing, the history holds only the current frame, which then fills both frame slots.
On HM3D-OVON, each episode fills \texttt{\textless{}instruction\textgreater{}} with \texttt{Walk to the \textless{}category\textgreater{} and stop by it.}, naming its goal category.
The answer lists sub-actions separated by commas, each of the form \texttt{forward \textless{}d\textgreater{} cm}, \texttt{turn left \textless{}a\textgreater{} degree}, \texttt{turn right \textless{}a\textgreater{} degree} or \texttt{stop}, and the first two are executed.

\subsection{Deviation Judge}
The judge is trained and deployed with the prompt below, where \texttt{\textless{}frame 1\textgreater{}}\,\dots\,\texttt{\textless{}frame 8\textgreater{}} are up to eight frames spaced evenly over the history before the current frame \texttt{\textless{}frame 9\textgreater{}}, \texttt{\textless{}actions\textgreater{}} lists up to the last twelve executed atomic actions or reads \texttt{(none)} before the first one, and \texttt{\textless{}chunk\textgreater{}} is the proposed chunk; actions are written as \texttt{MOVE\_FORWARD}, \texttt{TURN\_LEFT} or \texttt{TURN\_RIGHT} and separated by commas.
\begin{center}
\fbox{\begin{minipage}{0.95\linewidth}
\footnotesize\ttfamily\frenchspacing\raggedright\setlength{\parindent}{0pt}\setlength{\parskip}{1pt}
{\rmfamily\bfseries System}\par
You are a helpful assistant.\par
\smallskip
{\rmfamily\bfseries User}\par
\hangindent=1.5em You are a trajectory health critic for a vision-language navigation agent. Given the recent visual history, the instruction, and recent actions, judge whether the agent is on-track or off-track.\par
\hangindent=1.5em \mbox{}\ This is your historical observation:\textless{}frame 1\textgreater{}...\textless{}frame 8\textgreater{}\par
\hangindent=1.5em \mbox{}\ This is your current observation:\textless{}frame 9\textgreater{}\par
\hangindent=1.5em \mbox{}\ Your task is to \textless{}instruction\textgreater{}\par
\hangindent=1.5em \mbox{}\ Recent executed actions: [\textless{}actions\textgreater{}]\par
\hangindent=1.5em \mbox{}\ The base policy proposes the next action chunk: [\textless{}chunk\textgreater{}]. Is the agent off-track or losing progress? Answer one word from \{OK, lost, amb\}, where \textquotesingle{}OK\textquotesingle{} = making progress, \textquotesingle{}lost\textquotesingle{} = off-track or stagnating, \textquotesingle{}amb\textquotesingle{} = uncertain.\par
\smallskip
{\rmfamily\bfseries Assistant}\par
\textless{}label\textgreater{}
\end{minipage}}
\end{center}
Training supervises only the answer token, \texttt{lost} for LOST and \texttt{OK} for OK, and at deployment $m_t$ in \eqref{eq:score} is computed from the next-token probabilities of \texttt{lost} and \texttt{OK} at the answer position.

%% file: references.bib
@article{astranavmemory2026,
  title         = {{AstraNav-Memory}: Contexts Compression for Long Memory},
  author        = {Ren, Botao and Hu, Junjun and Xue, Xinda and Luo, Minghua and Chen, Jintao and Bai, Haochen and You, Liangliang and Xu, Mu},
  journal       = {arXiv preprint arXiv:2512.21627},
  year          = {2025},
  eprint        = {2512.21627},
  archivePrefix = {arXiv},
  primaryClass  = {cs.RO},
  doi           = {10.48550/arXiv.2512.21627},
  url           = {https://arxiv.org/abs/2512.21627}
}

@article{budvln2026,
  title         = {Nipping the Drift in the Bud: Retrospective Rectification for Robust Vision-Language Navigation},
  author        = {He, Gang and Liu, Zhenyang and Xu, Kepeng and Xu, Li and Qiao, Tong and Yu, Wenxin and Wu, Chang and Xie, Weiying},
  journal       = {arXiv preprint arXiv:2602.06356},
  year          = {2026},
  eprint        = {2602.06356},
  archivePrefix = {arXiv},
  primaryClass  = {cs.RO},
  doi           = {10.48550/arXiv.2602.06356},
  url           = {https://arxiv.org/abs/2602.06356}
}

@article{memoryadaptive2024,
  title   = {Memory-Adaptive Vision-and-Language Navigation},
  author  = {He, Keji and Jing, Ya and Huang, Yan and Lu, Zhihe and An, Dong and Wang, Liang},
  journal = {Pattern Recognition},
  volume  = {153},
  pages   = {110511},
  year    = {2024},
  doi     = {10.1016/j.patcog.2024.110511},
  url     = {https://doi.org/10.1016/j.patcog.2024.110511}
}

@inproceedings{anderson2018vln,
  title     = {Vision-and-Language Navigation: Interpreting Visually-Grounded Navigation Instructions in Real Environments},
  author    = {Anderson, Peter and Wu, Qi and Teney, Damien and Bruce, Jake and Johnson, Mark and S{\"u}nderhauf, Niko and Reid, Ian and Gould, Stephen and van den Hengel, Anton},
  booktitle = {Proceedings of the IEEE Conference on Computer Vision and Pattern Recognition (CVPR)},
  pages     = {3674--3683},
  year      = {2018},
  note      = {arXiv:1711.07280}
}

@inproceedings{krantz2020vlnce,
  title     = {Beyond the Nav-Graph: Vision-and-Language Navigation in Continuous Environments},
  author    = {Krantz, Jacob and Wijmans, Erik and Majumdar, Arjun and Batra, Dhruv and Lee, Stefan},
  booktitle = {European Conference on Computer Vision (ECCV)},
  pages     = {104--120},
  year      = {2020},
  note      = {arXiv:2004.02857}
}

@inproceedings{savva2019habitat,
  title     = {Habitat: A Platform for Embodied {AI} Research},
  author    = {Savva, Manolis and Kadian, Abhishek and Maksymets, Oleksandr and Zhao, Yili and Wijmans, Erik and Jain, Bhavana and Straub, Julian and Liu, Jia and Koltun, Vladlen and Malik, Jitendra and Parikh, Devi and Batra, Dhruv},
  booktitle = {Proceedings of the IEEE/CVF International Conference on Computer Vision (ICCV)},
  pages     = {9339--9347},
  year      = {2019},
  note      = {arXiv:1904.01201}
}

@inproceedings{zhang2024navid,
  title     = {{NaVid}: Video-based {VLM} Plans the Next Step for Vision-and-Language Navigation},
  author    = {Zhang, Jiazhao and Wang, Kunyu and Xu, Rongtao and Zhou, Gengze and Hong, Yicong and Fang, Xiaomeng and Wu, Qi and Zhang, Zhizheng and Wang, He},
  booktitle = {Robotics: Science and Systems (RSS)},
  year      = {2024},
  note      = {arXiv:2402.15852}
}

@inproceedings{cheng2024navila,
  title     = {{NaVILA}: Legged Robot Vision-Language-Action Model for Navigation},
  author    = {Cheng, An-Chieh and Ji, Yandong and Yang, Zhaojing and Gongye, Zaitian and Zou, Xueyan and Kautz, Jan and B{\i}y{\i}k, Erdem and Yin, Hongxu and Liu, Sifei and Wang, Xiaolong},
  booktitle = {Robotics: Science and Systems (RSS)},
  year      = {2025},
  note      = {arXiv:2412.04453}
}

@article{wei2025streamvln,
  title   = {{StreamVLN}: Streaming Vision-and-Language Navigation via {SlowFast} Context Modeling},
  author  = {Wei, Meng and Wan, Chenyang and Yu, Xiqian and Wang, Tai and Yang, Yuqiang and Mao, Xiaohan and Zhu, Chenming and Cai, Wenzhe and Wang, Hanqing and Chen, Yilun and Liu, Xihui and Pang, Jiangmiao},
  journal = {arXiv preprint arXiv:2507.05240},
  year    = {2025}
}

@inproceedings{yu2025correctnav,
  title     = {{CorrectNav}: Self-Correction Flywheel Empowers Vision-Language-Action Navigation Model},
  author    = {Yu, Zhuoyuan and Long, Yuxing and Yang, Zihan and Zeng, Chengyan and Fan, Hongwei and Zhang, Jiyao and Dong, Hao},
  booktitle = {Proceedings of the AAAI Conference on Artificial Intelligence (AAAI)},
  year      = {2026},
  note      = {arXiv:2508.10416}
}

@inproceedings{xin2026decovln,
  title     = {{DecoVLN}: Decoupling Observation, Reasoning, and Correction for Vision-and-Language Navigation},
  author    = {Xin, Zihao and Li, Wentong and Jiang, Yixuan and Wang, Bin and Cong, Runmin and Qin, Jie and Huang, Shengjun},
  booktitle = {Proceedings of the IEEE/CVF Conference on Computer Vision and Pattern Recognition (CVPR)},
  pages     = {32410--32420},
  year      = {2026},
  note      = {arXiv:2603.13133}
}

@inproceedings{navgrpo2026,
  title     = {Trajectory-Diversity-Driven Robust Vision-and-Language Navigation},
  author    = {Li, Jiangyang and Wan, Cong and Dong, Songlin and Ding, Chenhao and Wang, Qiang and Ma, Zhiheng and Gong, Yihong},
  booktitle = {Proceedings of the IEEE/CVF Conference on Computer Vision and Pattern Recognition (CVPR) Findings},
  pages     = {9143--9154},
  year      = {2026},
  note      = {arXiv:2603.15370}
}

@inproceedings{smartway2025,
  title     = {{SmartWay}: Enhanced Waypoint Prediction and Backtracking for Zero-Shot Vision-and-Language Navigation},
  author    = {Shi, Xiangyu and Li, Zerui and Lyu, Wenqi and Xia, Jiatong and Dayoub, Feras and Qiao, Yanyuan and Wu, Qi},
  booktitle = {IEEE/RSJ International Conference on Intelligent Robots and Systems (IROS)},
  year      = {2025},
  note      = {arXiv:2503.10069}
}

@article{sobolmark2026bpp,
  title   = {{BPP}: Long-Context Robot Imitation Learning by Focusing on Key History Frames},
  author  = {Sobol Mark, Max and Liang, Jacky and Attarian, Maria and Fu, Chuyuan and Dwibedi, Debidatta and Shah, Dhruv and Kumar, Aviral},
  journal = {arXiv preprint arXiv:2602.15010},
  year    = {2026}
}

@article{qwenrobotnav,
  title   = {{Qwen-RobotNav} Technical Report: A Scalable Navigation Model Designed for an Agentic Navigation System},
  author  = {Zhang, Jiazhao and Zhou, Gengze and Yin, Hale and Huang, Yiyang and Lei, Zixing and Peng, Qihang and Yuan, Haoqi and Zhang, Jie and Guo, Xudong and Chen, Xiaoyue and Yang, An and Huang, Fei and Yang, Zhibo and Lin, Junyang and Liu, Dayiheng and Zhou, Jingren and others},
  journal = {arXiv preprint arXiv:2606.18112},
  year    = {2026}
}

@article{futurenav,
  title   = {{FutureNav}: Unified World-Action Modeling for Vision-and-Language Navigation},
  author  = {Zhang, Lingfeng and Gong, Zeying and Hao, Xiaoshuai and Fu, Haoxiang and Zhang, Qiang and Zhou, Mingliang and Ye, Hangjun and Liang, Xiaojun and Liang, Junwei and Ding, Wenbo},
  journal = {arXiv preprint arXiv:2606.30367},
  year    = {2026}
}

@inproceedings{dehaan2019causal,
  title     = {Causal Confusion in Imitation Learning},
  author    = {de Haan, Pim and Jayaraman, Dinesh and Levine, Sergey},
  booktitle = {Advances in Neural Information Processing Systems (NeurIPS)},
  volume    = {32},
  year      = {2019},
  note      = {arXiv:1905.11979}
}

@inproceedings{wen2020fighting,
  title     = {Fighting Copycat Agents in Behavioral Cloning from Observation Histories},
  author    = {Wen, Chuan and Lin, Jierui and Darrell, Trevor and Jayaraman, Dinesh and Gao, Yang},
  booktitle = {Advances in Neural Information Processing Systems (NeurIPS)},
  volume    = {33},
  pages     = {2564--2575},
  year      = {2020},
  note      = {arXiv:2010.14876}
}

@inproceedings{ross2011reduction,
  title     = {A Reduction of Imitation Learning and Structured Prediction to No-Regret Online Learning},
  author    = {Ross, St{\'e}phane and Gordon, Geoffrey J. and Bagnell, J. Andrew},
  booktitle = {Proceedings of the Fourteenth International Conference on Artificial Intelligence and Statistics (AISTATS)},
  series    = {PMLR},
  volume    = {15},
  pages     = {627--635},
  year      = {2011},
  note      = {arXiv:1011.0686}
}

@article{zhu2026navida,
  title   = {{NaVIDA}: Vision-Language Navigation with Inverse Dynamics Augmentation},
  author  = {Zhu, Weiye and Zhang, Zekai and Wang, Xiangchen and Pan, Hewei
             and Wang, Teng and Geng, Tiantian and Xu, Rongtao and Zheng, Feng},
  journal = {arXiv preprint arXiv:2601.18188},
  year    = {2026}
}

@inproceedings{zhang2024uninavid,
  title     = {{Uni-NaVid}: A Video-based Vision-Language-Action Model for Unifying Embodied Navigation Tasks},
  author    = {Zhang, Jiazhao and Wang, Kunyu and Wang, Shaoan and Li, Minghan and Liu, Haoran and
               Wei, Songlin and Wang, Zhongyuan and Zhang, Zhizheng and Wang, He},
  booktitle = {Robotics: Science and Systems (RSS)},
  year      = {2025},
  doi       = {10.15607/RSS.2025.XXI.013},
  note      = {arXiv:2412.06224}
}

@book{vovk2005alrw,
  title     = {Algorithmic Learning in a Random World},
  author    = {Vovk, Vladimir and Gammerman, Alexander and Shafer, Glenn},
  publisher = {Springer},
  address   = {New York},
  year      = {2005},
  doi       = {10.1007/b106715},
}

@article{angelopoulos2021gentle,
  title   = {A Gentle Introduction to Conformal Prediction and Distribution-Free Uncertainty Quantification},
  author  = {Angelopoulos, Anastasios N. and Bates, Stephen},
  journal = {arXiv preprint arXiv:2107.07511},
  year    = {2021},
}

@inproceedings{hu2022lora,
  title     = {{LoRA}: Low-Rank Adaptation of Large Language Models},
  author    = {Hu, Edward J. and Shen, Yelong and Wallis, Phillip and Allen-Zhu, Zeyuan and
               Li, Yuanzhi and Wang, Shean and Wang, Lu and Chen, Weizhu},
  booktitle = {International Conference on Learning Representations (ICLR)},
  year      = {2022},
  note      = {arXiv:2106.09685}
}

@article{shao2024deepseekmath,
  title   = {{DeepSeekMath}: Pushing the Limits of Mathematical Reasoning in Open Language Models},
  author  = {Shao, Zhihong and Wang, Peiyi and Zhu, Qihao and Xu, Runxin and Song, Junxiao and
             Bi, Xiao and Zhang, Haowei and Zhang, Mingchuan and Li, Y. K. and Wu, Y. and Guo, Daya},
  journal = {arXiv preprint arXiv:2402.03300},
  year    = {2024},
}

@inproceedings{chen2021hamt,
  title     = {History Aware Multimodal Transformer for Vision-and-Language Navigation},
  author    = {Chen, Shizhe and Guhur, Pierre-Louis and Schmid, Cordelia and Laptev, Ivan},
  booktitle = {Advances in Neural Information Processing Systems (NeurIPS)},
  year      = {2021}
}

@inproceedings{ke2019tactical,
  title     = {Tactical Rewind: Self-Correction via Backtracking in Vision-and-Language Navigation},
  author    = {Ke, Liyiming and Li, Xiujun and Bisk, Yonatan and Holtzman, Ari and Gan, Zhe and Liu, Jingjing and Gao, Jianfeng and Choi, Yejin and Srinivasa, Siddhartha},
  booktitle = {IEEE Conference on Computer Vision and Pattern Recognition (CVPR)},
  year      = {2019}
}

@article{qi2025vlnr1,
  title   = {{VLN-R1}: Vision-Language Navigation via Reinforcement Fine-Tuning},
  author  = {Qi, Zhangyang and Zhang, Zhixiong and Yu, Yizhou and Wang, Jiaqi and Zhao, Hengshuang},
  journal = {arXiv preprint arXiv:2506.17221},
  year    = {2025}
}

@article{yue2025does,
  title   = {Does Reinforcement Learning Really Incentivize Reasoning Capacity in {LLMs} Beyond the Base Model?},
  author  = {Yue, Yang and Chen, Zhiqi and Lu, Rui and Zhao, Andrew and Wang, Zhaokai and Yue, Yang and Song, Shiji and Huang, Gao},
  journal = {arXiv preprint arXiv:2504.13837},
  year    = {2025},
  note    = {NeurIPS 2025}
}

@inproceedings{yokoyama2024hm3dovon,
  title     = {{HM3D-OVON}: A Dataset and Benchmark for Open-Vocabulary Object Goal Navigation},
  author    = {Yokoyama, Naoki and Ramrakhya, Ram and Das, Abhishek and Batra, Dhruv and Ha, Sehoon},
  booktitle = {IEEE/RSJ International Conference on Intelligent Robots and Systems (IROS)},
  year      = {2024},
  note      = {arXiv:2409.14296}
}

@article{richards2017persistence,
  title   = {The Persistence and Transience of Memory},
  author  = {Richards, Blake A. and Frankland, Paul W.},
  journal = {Neuron},
  volume  = {94},
  number  = {6},
  pages   = {1071--1084},
  year    = {2017},
  doi     = {10.1016/j.neuron.2017.04.037}
}

@article{zacks2007event,
  title   = {Event Perception: A Mind--Brain Perspective},
  author  = {Zacks, Jeffrey M. and Speer, Nicole K. and Swallow, Khena M. and Braver, Todd S. and Reynolds, Jeremy R.},
  journal = {Psychological Bulletin},
  volume  = {133},
  number  = {2},
  pages   = {273--293},
  year    = {2007},
  doi     = {10.1037/0033-2909.133.2.273}
}

@article{schulman2017proximal,
  title   = {Proximal Policy Optimization Algorithms},
  author  = {Schulman, John and Wolski, Filip and Dhariwal, Prafulla and Radford, Alec and Klimov, Oleg},
  journal = {arXiv preprint arXiv:1707.06347},
  year    = {2017}
}

@article{anderson2018evaluation,
  title   = {On Evaluation of Embodied Navigation Agents},
  author  = {Anderson, Peter and Chang, Angel and Chaplot, Devendra Singh and Dosovitskiy, Alexey and Gupta, Saurabh and Koltun, Vladlen and Kosecka, Jana and Malik, Jitendra and Mottaghi, Roozbeh and Savva, Manolis and Zamir, Amir R.},
  journal = {arXiv preprint arXiv:1807.06757},
  year    = {2018}
}

@article{jeong2026navheads,
  title         = {Your Vision-Language-Action Model Already Has Attention Heads For Path Deviation Detection},
  author        = {Jeong, Jaehwan and Zhu, Evelyn and Lin, Jinying and Jaimes, Emmanuel and Vu, Tuan-Anh and Joo, Jungseock and Kim, Sangpil and Jawed, M. Khalid},
  journal       = {arXiv preprint arXiv:2603.13782},
  year          = {2026},
  eprint        = {2603.13782},
  archivePrefix = {arXiv},
  primaryClass  = {cs.RO},
  url           = {https://arxiv.org/abs/2603.13782}
}

@inproceedings{codevilla2019exploring,
  title     = {Exploring the Limitations of Behavior Cloning for Autonomous Driving},
  author    = {Codevilla, Felipe and Santana, Eder and L{\'o}pez, Antonio M. and Gaidon, Adrien},
  booktitle = {Proceedings of the IEEE/CVF International Conference on Computer Vision (ICCV)},
  pages     = {9328--9337},
  year      = {2019},
  doi       = {10.1109/ICCV.2019.00942},
  note      = {arXiv:1904.08980}
}

@inproceedings{ma2019selfmonitoring,
  title     = {Self-Monitoring Navigation Agent via Auxiliary Progress Estimation},
  author    = {Ma, Chih-Yao and Lu, Jiasen and Wu, Zuxuan and AlRegib, Ghassan and Kira, Zsolt and Socher, Richard and Xiong, Caiming},
  booktitle = {International Conference on Learning Representations (ICLR)},
  year      = {2019},
  url       = {https://openreview.net/forum?id=r1GAsjC5Fm},
  note      = {arXiv:1901.03035}
}

@inproceedings{ma2019regretful,
  title     = {The Regretful Agent: Heuristic-Aided Navigation Through Progress Estimation},
  author    = {Ma, Chih-Yao and Wu, Zuxuan and AlRegib, Ghassan and Xiong, Caiming and Kira, Zsolt},
  booktitle = {Proceedings of the IEEE/CVF Conference on Computer Vision and Pattern Recognition (CVPR)},
  pages     = {6725--6733},
  year      = {2019},
  doi       = {10.1109/CVPR.2019.00689},
  note      = {arXiv:1903.01602}
}

@article{liufu2026repovla,
  title         = {{RePO-VLA}: Recovery-Driven Policy Optimization for Vision-Language-Action Models},
  author        = {Liufu, Weijia and Guo, Xiaoyu and Chen, Ruiyi and Liu, Jingzhi and Zhang, Kaidong and Liang, Xiwen and Lin, Jianqi and Sun, Dawei and Wang, Yuze and Xu, Rongtao and Lin, Bingqian and Yang, Bowen and Cao, Tongtong and Peng, Bowen and Zhang, Dongyu and Wang, Guangrun and Wang, Min and Lin, Liang and Liang, Xiaodan},
  journal       = {arXiv preprint arXiv:2605.09410},
  year          = {2026},
  eprint        = {2605.09410},
  archivePrefix = {arXiv},
  primaryClass  = {cs.RO},
  url           = {https://arxiv.org/abs/2605.09410}
}

@inproceedings{krantz2021waypoint,
  title     = {Waypoint Models for Instruction-guided Navigation in Continuous Environments},
  author    = {Krantz, Jacob and Gokaslan, Aaron and Batra, Dhruv and Lee, Stefan and Maksymets, Oleksandr},
  booktitle = {Proceedings of the IEEE/CVF International Conference on Computer Vision (ICCV)},
  year      = {2021},
  note      = {arXiv:2110.02207}
}

@inproceedings{wang2023scalevln,
  title     = {Scaling Data Generation in Vision-and-Language Navigation},
  author    = {Wang, Zun and Li, Jialu and Hong, Yicong and Wang, Yi and Wu, Qi and Bansal, Mohit and Gould, Stephen and Tan, Hao and Qiao, Yu},
  booktitle = {Proceedings of the IEEE/CVF International Conference on Computer Vision (ICCV)},
  pages     = {11975--11986},
  year      = {2023},
  doi       = {10.1109/ICCV51070.2023.01103},
  note      = {arXiv:2307.15644}
}

@inproceedings{hong2022bridging,
  title     = {Bridging the Gap Between Learning in Discrete and Continuous Environments for Vision-and-Language Navigation},
  author    = {Hong, Yicong and Wang, Zun and Wu, Qi and Gould, Stephen},
  booktitle = {Proceedings of the IEEE/CVF Conference on Computer Vision and Pattern Recognition (CVPR)},
  year      = {2022},
  doi       = {10.1109/CVPR52688.2022.01500},
  note      = {arXiv:2203.02764}
}

@article{an2024etpnav,
  title   = {{ETPNav}: Evolving Topological Planning for Vision-Language Navigation in Continuous Environments},
  author  = {An, Dong and Wang, Hanqing and Wang, Wenguan and Wang, Zun and Huang, Yan and He, Keji and Wang, Liang},
  journal = {IEEE Transactions on Pattern Analysis and Machine Intelligence (TPAMI)},
  year    = {2024},
  doi     = {10.1109/TPAMI.2024.3386695},
  note    = {arXiv:2304.03047}
}

@inproceedings{yao2025navmorph,
  title     = {{NavMorph}: A Self-Evolving World Model for Vision-and-Language Navigation in Continuous Environments},
  author    = {Yao, Xuan and Gao, Junyu and Xu, Changsheng},
  booktitle = {Proceedings of the IEEE/CVF International Conference on Computer Vision (ICCV)},
  year      = {2025},
  note      = {arXiv:2506.23468}
}

@inproceedings{zhang2026navfom,
  title     = {Embodied Navigation Foundation Model},
  author    = {Zhang, Jiazhao and Li, Anqi and Qi, Yunpeng and Li, Minghan and Liu, Jiahang and Wang, Shaoan and Liu, Haoran and Zhou, Gengze and Wu, Yuze and Li, Xingxing and Fan, Yuxin and Li, Wenjun and Chen, Zhibo and Gao, Fei and Wu, Qi and Zhang, Zhizheng and Wang, He},
  booktitle = {International Conference on Learning Representations (ICLR)},
  year      = {2026},
  note      = {arXiv:2509.12129}
}

@inproceedings{zeng2026janusvln,
  title     = {{JanusVLN}: Decoupling Semantics and Spatiality with Dual Implicit Memory for Vision-Language Navigation},
  author    = {Zeng, Shuang and Qi, Dekang and Chang, Xinyuan and Xiong, Feng and Xie, Shichao and Wu, Xiaolong and Liang, Shiyi and Xu, Mu and Wei, Xing and Guo, Ning},
  booktitle = {International Conference on Learning Representations (ICLR)},
  year      = {2026},
  note      = {arXiv:2509.22548}
}

@inproceedings{wei2026dualvln,
  title     = {Ground Slow, Move Fast: A Dual-System Foundation Model for Generalizable Vision-and-Language Navigation},
  author    = {Wei, Meng and Wan, Chenyang and Peng, Jiaqi and Yu, Xiqian and Yang, Yuqiang and Feng, Delin and Cai, Wenzhe and Zhu, Chenming and Wang, Tai and Pang, Jiangmiao and Liu, Xihui},
  booktitle = {International Conference on Learning Representations (ICLR)},
  year      = {2026},
  note      = {arXiv:2512.08186}
}

@inproceedings{yao2026sc2wm,
  title     = {{SC$^{2}$-WM}: A Self-Correcting World Model with Closed-Loop Feedback for Vision-and-Language Navigation in Continuous Environments},
  author    = {Yao, Xuan and Zhu, Yuze and Gao, Junyu and Wang, Zongmeng and Xu, Changsheng},
  booktitle = {International Conference on Machine Learning (ICML)},
  year      = {2026},
  note      = {arXiv:2608.07548}
}

@article{li2026saca,
  title   = {Let's Reward Step-by-Step: Step-Aware Contrastive Alignment for Vision-Language Navigation in Continuous Environments},
  author  = {Li, Haoyuan and Liu, Rui and Fan, Hehe and Yang, Yi},
  journal = {arXiv preprint arXiv:2603.09740},
  year    = {2026}
}

@article{wang2026reflectvln,
  title   = {{ReflectVLN}: Training Vision-Language Navigation Agents with Reflective Reasoning},
  author  = {Wang, Jiahang and Yang, Yirong and Zhu, Yanqing and Luo, Minghua and Xie, Shichao and Liu, Fei and Xu, Mu},
  journal = {arXiv preprint arXiv:2607.12680},
  year    = {2026}
}

@inproceedings{yokoyama2024vlfm,
  title     = {{VLFM}: Vision-Language Frontier Maps for Zero-Shot Semantic Navigation},
  author    = {Yokoyama, Naoki and Ha, Sehoon and Batra, Dhruv and Wang, Jiuguang and Bucher, Bernadette},
  booktitle = {IEEE International Conference on Robotics and Automation (ICRA)},
  year      = {2024},
  note      = {arXiv:2312.03275}
}

@inproceedings{zhu2025mtu3d,
  title     = {Move to Understand a {3D} Scene: Bridging Visual Grounding and Exploration for Efficient and Versatile Embodied Navigation},
  author    = {Zhu, Ziyu and Wang, Xilin and Li, Yixuan and Zhang, Zhuofan and Ma, Xiaojian and Chen, Yixin and Jia, Baoxiong and Liang, Wei and Yu, Qian and Deng, Zhidong and Huang, Siyuan and Li, Qing},
  booktitle = {IEEE/CVF International Conference on Computer Vision (ICCV)},
  year      = {2025},
  note      = {arXiv:2507.04047}
}

@article{chu2026abotn0,
  title   = {{ABot-N0}: Technical Report on the {VLA} Foundation Model for Versatile Embodied Navigation},
  author  = {Chu, Zedong and Xie, Shichao and Wu, Xiaolong and Shen, Yanfen and Luo, Minghua and Wang, Zhengbo and Liu, Fei and Leng, Xiaoxu and Hu, Junjun and Yin, Mingyang and others},
  journal = {arXiv preprint arXiv:2602.11598},
  year    = {2026}
}
